\documentclass[sigconf]{acmart}

\def\BibTeX{{\rm B\kern-.05em{\sc i\kern-.025em b}\kern-.08emT\kern-.1667em\lower.7ex\hbox{E}\kern-.125emX}}
    
\copyrightyear{2019} 
\acmYear{2019} 
\setcopyright{acmlicensed}
\acmConference[MM '19] {Proceedings of the 27th ACM International Conference on Multimedia}{October 21--25, 2019}{Nice, France}
\acmBooktitle{Proceedings of the 27th ACM Int'l Conf. on Multimedia (MM'19), Oct. 21--25, 2019, Nice, France}
\acmPrice{15.00}
\acmDOI{10.1145/3343031.3351020}
\acmISBN{978-1-4503-6889-6/19/10}

\usepackage[utf8]{inputenc}
\usepackage{breqn}
\usepackage{gensymb}
\usepackage{multirow}
\usepackage{graphicx}
\usepackage{todonotes}
\usepackage{enumitem}
\usepackage{flushend}
\usepackage{hyperref}

\setlist{leftmargin=0mm}
\usepackage{array}
\newcommand{\ra}[1]{\renewcommand{\arraystretch}{#1}}

\usepackage{relsize}
\usepackage{textcomp}
\usepackage{tabularx}

\usepackage{listings}

\usepackage[position=top, labelformat=empty, singlelinecheck=false, justification=centering,farskip=1.8pt]{subfig}
\def \qwidth {0.12}
\def \qwidthL {0.137}
\def \qwidthA {0.192}
\def \qwidthSix {0.16}

\usepackage{tikz}
\usetikzlibrary{shapes.geometric}
\newcommand*{\squarecat}[1]{\begin{tikzpicture}[every node/.style={draw=none,rectangle,inner sep=0pt}]
		\node[minimum size=\qwidthL\textwidth] (0,0)  { #1};%
		\end{tikzpicture}}

\usepackage[font=small,belowskip=-7pt,aboveskip=0pt]{caption}

\usepackage{balance}

\usepackage[scaled]{beramono}
\usepackage{listings}

\usepackage{color}
\definecolor{Orange}{rgb}{0.9,0.5,0}
\definecolor{NavyBlue}{rgb}{0.1, 0.4, 0.8}
\definecolor{Magenta}{rgb}{0.8, 0.1, 0.6}
\definecolor{Red}{rgb}{1, 0, 0}

\newcommand{\TGAN}{${\mathlarger{\mathlarger{\bigtriangleup}}}$-{GAN}}
\newcommand{\boldTGAN}{$\pmb{\mathlarger{\mathlarger{\bigtriangleup}}}$-{GAN}}
\usepackage[belowskip=-0.3em,aboveskip=2pt]{caption}
\begin{document}
\fancyhead{}
%
\title{Gesture-to-Gesture Translation in the Wild via Category-Independent Conditional Maps}
%

\author{Yahui Liu}
\email{yahui.liu@unitn.it}
\affiliation[obeypunctuation=true]{\institution{University of Trento}, \city{Trento}, \country{Italy}}

\author{Marco De Nadai}
\email{denadai@fbk.eu}
\orcid{0000-0001-8466-3933}
\affiliation[obeypunctuation=true]{\institution{FBK}, \city{Trento}, \country{Italy}}

\author{Gloria Zen}
\email{gloria.zen@unitn.it}
\affiliation[obeypunctuation=true]{\institution{University of Trento}, \city{Trento}, \country{Italy}}

\author{Nicu Sebe}
\email{niculae.sebe@unitn.it}
\affiliation[obeypunctuation=true]{\institution{University of Trento}, \city{Trento}, \country{Italy}}

\author{Bruno Lepri}
\email{lepri@fbk.eu}
\affiliation[obeypunctuation=true]{\institution{FBK}, \city{Trento}, \country{Italy}}


%
\renewcommand{\shortauthors}{Liu, et al.}
\renewcommand{\shorttitle}{Gesture-to-Gesture Translation in the Wild}

%
\begin{abstract}
\begin{sloppypar}
Recent works have shown Generative Adversarial Networks (GANs) to be particularly effective in image-to-image translations.
However, in tasks such as body pose and hand gesture translation, existing methods usually require precise annotations, e.g. key-points or skeletons, which are time-consuming to draw. 
In this work, we propose a novel GAN architecture that decouples the required annotations into a category label - that specifies the gesture type - and a simple-to-draw category-independent conditional map - that expresses the location, rotation and size of the hand gesture.
Our architecture synthesizes the target gesture while preserving the background context, thus effectively dealing with gesture translation \textit{in the wild}.
To this aim, we use an attention module and a rolling guidance approach, which loops the generated images back into the network and produces higher quality images compared to competing works.
Thus, our GAN learns to generate new images from simple annotations without requiring key-points or skeleton labels.
Results on two public datasets show that our method outperforms state of the art approaches both quantitatively and qualitatively.
To the best of our knowledge, no work so far has addressed the gesture-to-gesture translation \emph{in the wild} by requiring user-friendly annotations.

\end{sloppypar}

\end{abstract}

%
%


%

%

%
\maketitle

\begin{figure}[t]
    \includegraphics[width=1\columnwidth]{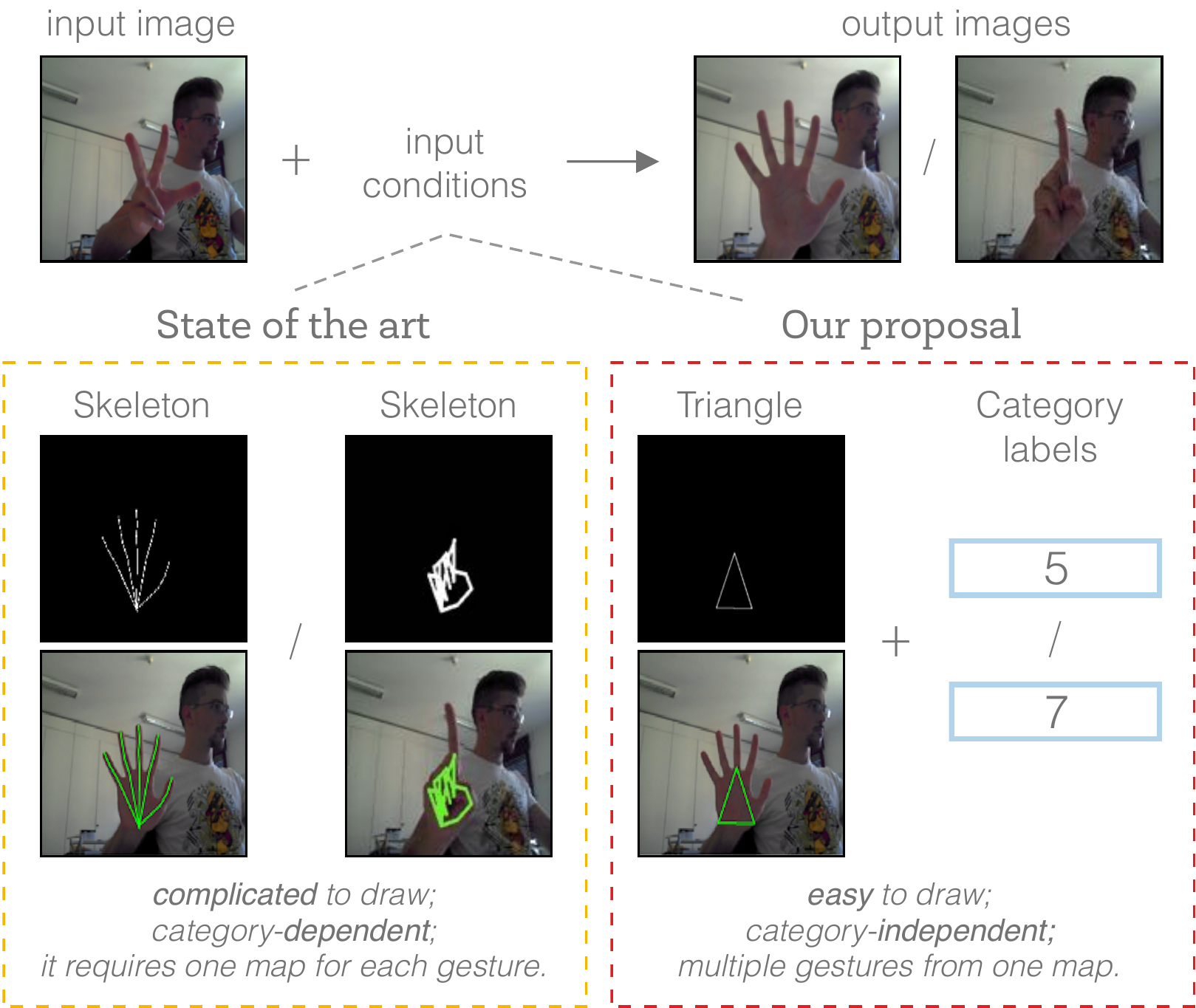}
    \caption{
    Our proposal decouples the \emph{category label} that specifies the gesture type (e.g., gesture "5" or "7") from the \emph{conditional map} (i.e. triangle) that controls the location, orientation and size of the target gesture.
    Existing works require a detailed conditional map (e.g. skeleton) that is gesture-dependent. 
    In this example, we show that our method significantly lowers the drawing effort and expertise required by users. Our method can generate multiple output images with the same map for multiple gesture categories. 
    }
    \label{fig:teaser}
    \vspace{-0.3em}
\end{figure}

\section{Introduction}
\begin{sloppypar}
Photo-editing software, fashion, and retail markets would enormously benefit from the possibility of modifying an image through a simple user input describing the changes to make (e.g. change the hand gesture of the person in the picture from ``open hand'' to ``ok'').
However, despite the significant advances of Generative Adversarial Networks (GANs)~\cite{goodfellow2014generative,radford2015unsupervised,arjovsky2017wasserstein,gulrajani2017improved}, the generation of images \emph{in the wild} without precise annotations (e.g. hand skeleton) is still an open problem.
Previous literature on image-to-image translation has relied either on pixel-to-pixel mappings~\cite{isola2017image,zhu2017unpaired,yang2018crossing} or precise annotations to localize the instance to be manipulated, such as segmentation masks~\cite{mo2018instagan}, key-points~\cite{siarohin2018deformable}, skeletons~\cite{tang2018gesturegan} and facial landmarks~\cite{sanchez2018triple}. 
However, obtaining such annotations is not trivial.
On one hand, automatic methods for key-points extraction~\cite{cao2017realtime,simon2017hand} may fail or the reference gesture image/video~\cite{siarohin2018animating} may be not available. On the other, drawing such annotations is complicated, time-consuming, and their quality directly affects the performance of the network. 

Moreover, existing methods often focus on foreground content, e.g., the target gesture or facial expression, generating blurred and imprecise backgrounds~\cite{siarohin2018deformable,ma2017pose,reed2016learning}.
These methods are well suited in the cases where
share fixed or similar spatial structures, such as facial expression datasets~\cite{liu2015deep,fabian2016emotionet}.
Instead, in image-to-image translations \textit{in the wild}, both the foreground and background between the source image and target image can vary a lot~\cite{tang2018gesturegan}. 

In this paper, we propose a novel method, named as \TGAN, that requires a simple-to-draw annotation, such as a triangle, and focuses on the challenging task of hand gesture-to-gesture translation \emph{in the wild}.
In general, annotations such as key-points or skeletons are category-dependent since they provide four types of information at the same time, namely \textit{category} (e.g., gesture ``5"), \textit{location}, \textit{scale} and \textit{orientation} of the hand gesture.
Instead, our \TGAN~decouples the \textit{category} from the \textit{location}-\textit{scale}-\textit{orientation} information. 
Using a category-independent conditional map significantly lowers the annotation cost, allowing to generate multiple target images while requiring users to draw only a single map. 
In this work, we refer to ``annotations'' as the user effort to draw the desired target gesture also at deploy time, besides the effort needed for generating the training data for our model. 
The intuition of our approach is depicted in Figure~\ref{fig:teaser}.
Furthermore, we propose a novel architecture that uses an attention module and a rolling guidance approach to perform gesture-to-gesture translation \textit{in the wild}.
Our research yields three main contributions:
\begin{itemize}
    \item \emph{Decoupled conditional map and category.} We designed a general architecture for gesture-to-gesture translations that separately encodes the category label (e.g., gesture ``5") and the category-independent conditional map. 
    This allows to perform several image translations with the same conditional map.
    \item \emph{Rolling guidance and attention mask.} We propose a novel rolling guidance approach that allows generating higher quality output images by feeding the generated image back to the input as an additional condition to be considered. 
    Also, \TGAN~learns unsupervisedly an attention mask that allows to preserve the details shared between the input and target image. 
    \item \emph{Simple-to-draw conditional maps.} 
    We propose a triangle conditional map as the simplest and minimal necessary user provided condition for gesture-to-gesture translation.
    To the best of our knowledge, no work so far has addressed the gesture-to-gesture translation task \emph{in the wild} by requiring user-friendly annotations. 
    Furthermore, we assess the performance of our method with different shapes, such as boundary and skeleton maps.
    Finally, we enrich two public datasets with different conditional maps for each gesture image, specifically based on triangles and boundaries.
\end{itemize}
\end{sloppypar}

\begin{figure*}[ht!]
    \centering
    \includegraphics[width=0.95\linewidth]{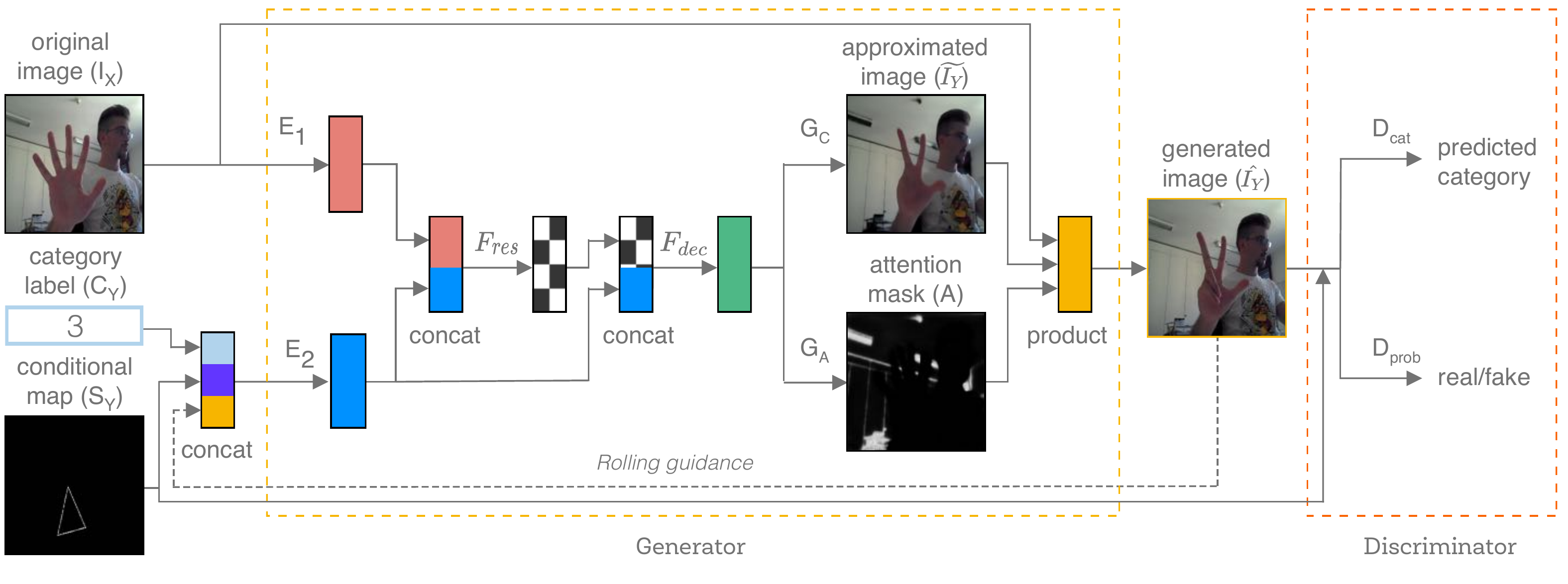}
    \vspace{-1em}
    \caption{Overview of \boldTGAN, that allows to translate hand gestures \emph{in the wild} by separately encoding the input image and the target gesture attributes including the category label (e.g., gesture 3) and the category-independent conditional map. We include an unsupervised attention mask to preserve the details shared between the input image and the target image. Specially, we feed the first reconstructed image back to the input conditions encoding module to improve quality of output image.}
    \label{fig:ourGAN}
\end{figure*}

\section{Related work}
Recently, there have been a lot of works on Generative Adversarial Networks (GANs) and, particularly, on conditional GANs for the task of image-to-image translation~\cite{isola2017image,zhu2017unpaired,zhu2017toward,choi2018stargan,pumarola2018ganimation}. 
A significant line of these works have focused on translation tasks where the the input and target images are spatially aligned, as in the case of style transfer~\cite{zhu2017unpaired,johnson2016perceptual,isola2017image}, emotional content transfer~\cite{emotionGAN18} or 
image inpaiting~\cite{zhang2018semantic}.
In general, these works aim to preserve the main image content while presenting them in various styles.
Another line of works have tackled the more challenging task of image translation where the target object is spatially unaligned with respect to the original location, shape or size of the input original object.
This is the case of image-to-image translation tasks like facial expression~\cite{sanchez2018triple,geng20193d}, 
human pose~\cite{ma2017pose,siarohin2018deformable} or 
hand gesture translation~\cite{tang2018gesturegan}.
To this aim, methods usually require geometry information as guidance to guarantee where and how to edit visual patterns corresponding to the image attributes,
such as key-points~\cite{ma2017pose,siarohin2018animating,ma2018disentangled}, skeletons~\cite{tang2018gesturegan}, object segmentation masks~\cite{mo2018instagan}, facial landmarks~\cite{sanchez2018triple}, action units~\cite{pumarola2018ganimation} or 3D models~\cite{geng20193d}. 
 GANimation~\cite{pumarola2018ganimation} learns an attention mask to preserve the background content of the source image for the spatially aligned translation of facial expressions. 
 InstaGAN~\cite{mo2018instagan} performs multi-instance domain-to-domain image translation by requiring as input precise segmentation masks for the target objects.
 Pose Guided Person Generation Network (PG$^2$)~\cite{ma2017pose} proposes a two stages generating framework to refine the output image given a reference image and a target pose.
 MonkeyNet~\cite{siarohin2018animating} generates a conditioned video transferring body movements from a target driving video to a reference appearance image. The target key-points are obtained automatically using state-of-the-art detectors.
Relatively few works have considered the challenging task of image translation \textit{in the wild}, where both the foreground content and the background context undergoes significant variation~\cite{tang2018gesturegan}. 
In these cases, not only the networks learn to synthesize the target object or attribute, but they also have to correctly locate it in the image while the pixels depicting the content are preserved from the original image.
GestureGAN~\cite{tang2018gesturegan} proposes a novel color loss to improve the output image quality and to produce sharper results. However, this approach does not aim to separately encode the foreground content and the background context.
Furthermore, the large majority of image-to-image translations works focus on one to one domain mapping. 
Recently, efficient solutions have been proposed to address the multi-domain translations~\cite{choi2018stargan,geng20193d}. In particular, StarGAN~\cite{choi2018stargan} proposes the use of multiple mask vectors to generalize on different datasets and domain labels. 3DMM~\cite{geng20193d} decomposes an image into shape and texture spaces, and relies on an identity and target coefficients to generate images in multiple domains. They, however, focus on multi-domain face attributes where neither the background quality or the misalignment are considered.

Our work falls within the category of multi-domain and image-to-image translation \emph{in the wild}. Still, rather than asking expensive annotation to the user, we propose a novel user-friendly annotation strategy.
To the best of our knowledge, no works so far have investigated other approaches for gesture translation that also allow to reduce the annotation effort.


\section{Our approach}
The goal of this work is to perform gesture translation \emph{in the wild}, conditioned on the user provided gesture attributes, such as the desired category, location, scale and orientation. 
Specifically, we seek to estimate the mapping $\mathcal{M}$: $(\mathbf{I}_X, \mathbf{S}_Y, C_Y) \rightarrow \mathbf{I}_Y$ that translates an input image $\mathbf{I}_X$ into an output image $\mathbf{I}_Y$, conditioned to a hand gesture category $C_Y$ and a simple-to-draw conditional map $\mathbf{S}_Y$, which encodes the desired location, scale and orientation of the gesture.
The generated image is discriminated through a Conditional Critic discriminator~\cite{pumarola2018ganimation} that judges the photo-realism of the generated image and assess the category of the generated gesture (e.g. gesture "5"). 
Furthermore, in order to efficiently deal with the gesture translation \emph{in the wild}, we propose a rolling guidance approach and an attention mask module. 
Figure~\ref{fig:ourGAN} depicts the architecture of our approach. In this Section we further explain the details of the architecture and the loss functions used to train our framework.

\subsection{Network Architecture}
\label{sec:network}


\noindent \textbf{Generator.}  
Our Generator $G$ takes as input the conditional image $\textbf{I}_X$, the category label $C_Y$ and the conditional map $\textbf{S}_Y$
and it outputs the translated image $\textbf{I}_Y$.
The original image and the target gesture conditions are encoded 
respectively through the encoder $E_1$ and $E_2$, and then concatenated and provided to $F_{res}$. Then, the output features of $E_2$ and $F_{res}$ are concatenated and provided to $F_{dec}$. 
Similarly to~\cite{pumarola2018ganimation}, in the decoder we learn to predict in an unsupervised manner the approximated image $\widetilde{\textbf{I}}_Y$ and an \textit{Attention Mask} $\mathbf{A} \in [0,1.0]^{{H\times W}}$. 
Since both the background and foreground between the input and target images can vary a lot, the learned attention mask $\mathbf{A}$ tends to preserve the shared part between $\mathbf{I}_X$ and $\mathbf{I}_Y$. 
Pixels in $\mathbf{A}$ with higher value indicate to preserve the corresponding pixels from $\mathbf{I}_X$, otherwise from $\widetilde{\mathbf{I}}_Y$.
The final generated image is thus obtained as:
\begin{equation}
\widehat{\mathbf{I}}_Y = \mathbf{A}*\mathbf{I}_X + (1-\mathbf{A})*\widetilde{\mathbf{I}}_Y
\label{eq:attention-mask}
\end{equation}
Furthermore, the generated image $\mathbf{I}_Y$ is rolled back as an input condition and concatenated together with the category label $C_Y$ and the conditional map $\textbf{S}_Y$. More details are provided in Section~\ref{Sec:rolling-guidance}.  

\medskip
\noindent \textbf{Discriminator.} Our Discriminator $D$ takes as input the generated image $\widehat{\textbf{I}}_Y$ and its conditional map $\textbf{S}_Y$. 
The outputs of $D$ consist of two parts: $D_{cat}$ predicts the category label 
and $D_{prob}$ classifies whether local image patches are real or fake. 
As in~\cite{pumarola2018ganimation}, D$_{prob}$ is based on PatchGANs~\cite{isola2017image}.
The conditional map $\textbf{S}_Y$ is needed by $D$ to verify that the generated gesture has also the right location, scale and orientation.

We refer to the Supplementary Material for additional details in the architecture.

\subsection{Objective Formulation}
\label{sec:loss}

The loss function of \TGAN\ is composed of four main components, namely \emph{GAN Loss} that pushes the generated image distribution to the distribution of source images; \emph{Reconstruction loss} that forces the generator to reconstruct the source and target image; \emph{Category Label loss} that allows to properly classify the generated image into hand gesture classes; and \emph{Total Variation loss} that indirectly learns the attention mask in an unsupervised fashion.

\medskip
\noindent \textbf{GAN Loss.} To generate images with the same distribution of the source images, 
we adopt an adversarial loss: 
\begin{equation}
\begin{aligned}
\mathcal{L}_{GAN} = & \mathbb{E}_{\mathbf{I}_Y, \mathbf{S}_Y\sim \mathbb{P}}[\log D_{prob}(\mathbf{I}_Y, \mathbf{S}_Y)] + \\ & \mathbb{E}_{\mathbf{I}_X,\mathbf{S}_Y, C_Y\sim \mathbb{P}}[\log (1- D_{prob}(G(\mathbf{I}_X, \mathbf{S}_Y, C_Y), \mathbf{S}_Y))] \\
\end{aligned}
\end{equation}
%
%
where $\mathbb{P}$ is the data distribution of the hand gesture images in the dataset, and 
where $G$ generates an image $G(\mathbf{I}_X, \mathbf{S}_Y, C_Y)$ conditioned on both the input image $\mathbf{I}_X$, the conditional map $\mathbf{S}_Y$, and the target category $C_Y$, while $D$ tries to distinguish between real and fake images.  In this paper, we refer to the term $D_{prob}(\mathbf{I}, \mathbf{S})$ as a probability distribution over sources given by $D$.

\medskip
\noindent \textbf{Reconstruction Loss.} The adversarial loss does not guarantee that the generated image is consistent with both the target conditional map $\mathbf{S}$ and category $C$.
Thus, we first apply a \emph{forward reconstruction loss} that ties together the target image $\mathbf{I}_Y$ with its target conditional map $\mathbf{S}_Y$ and category $C_Y$:
\begin{equation}
    \mathcal{L}_{rec} = \|G(\mathbf{I}_X, \mathbf{S}_Y, C_Y) - \mathbf{I}_Y\|_1
\end{equation}
%
Then, instead of using perceptual features (e.g. extracted from VGG~\cite{very2015simonyan} networks) to force the model to reconstruct the source image, we propose a simplified  \emph{self-reconstruction (identity) loss}:
\begin{equation}
    \mathcal{L}_{idt} = \|G(\mathbf{I}_X, \mathbf{S}_X, C_X) - \mathbf{I}_X \|_1
\end{equation}
where $\mathbf{S}_X$ is the conditional map of the source image and $C_X$ the category label of the source image.
Finally, we apply the \emph{cycle consistency loss}~\cite{zhu2017unpaired,kim2017learning} to reconstruct the original image from the generated one:
\begin{equation}
    \mathcal{L}_{cyc}=\|G(G(\mathbf{I}_X, \mathbf{S}_Y, C_Y), \mathbf{S}_X, C_X) - \mathbf{I}_X\|_1
\end{equation}
Note that we apply the cycle reconstruction loss only in one direction to reduce computation, i.e., A-to-B-to-A, since a translation pair based on two images A and B may be sampled either as A-to-B or as B-to-A
during the training. 

\medskip
\noindent \textbf{Category Label loss.} We enforce the generator to render realistic samples that have to be correctly classified to the hand gesture expressed by the input category label.
Similarly to StarGAN~\cite{choi2018stargan}, we split the \emph{Category Label loss} in two terms: a gesture classification loss of the real image $\mathbf{I}_Y$ used to optimize $D$, and a gesture classification loss of the generated image $\hat{\mathbf{I}}_Y$, used to optimize $G$.
Specifically:
\begin{equation}
    \mathcal{L}_{cls} = \mathbb{E}_{\mathbf{I}_Y, C_Y}[- \log D_{cat}(C_Y | \mathbf{I}_Y, \mathbf{S}_Y)]
\end{equation}
\begin{equation}
\mathcal{L}_{\hat{cls}} = \mathbb{E}_{\hat{\mathbf{I}}_Y, C_Y}[- \log D_{cat}(C_Y | \hat{\mathbf{I}}_Y, \mathbf{S}_Y)]
\end{equation}
where $D_{cat}(C_Y | \mathbf{I}_Y, \mathbf{S}_Y)$ and $D_{cat}(C_Y | \hat{\mathbf{I}}_Y,\mathbf{S}_Y)$ represent a probability distribution over the categories of hand gestures respectively in the real and generated images.
In other words, these losses allow to generate images that can be correctly classified as the target hand gesture category.

\medskip
\noindent \textbf{Total Variation loss}. 
To prevent the final generated image having artifacts, we use a Total Variation Regularization, $f_{tv}$,
as in GANimation~\cite{pumarola2018ganimation}. 
However, differently from them, we calculate $f_{tv}$ over the approximated image $\widetilde{\mathbf{I}}_Y$ instead of the attention mask $\mathbf{A}$, 
thus allowing to freely explore the shared pixels between the source and target images. 
The total variation loss is applied both to the \textit{forward reconstruction} and \textit{self-reconstruction} and is formulated as:
\begin{equation}
    \mathcal{L}_{tv} = f_{tv}(G_C(\mathbf{I}_X, \mathbf{S}_X, C_X)+f_{tv}(G_C(\mathbf{I}_X, \mathbf{S}_Y, C_Y))
\end{equation}
The total variation regularization $f_{tv}$ is defined as:
\begin{dmath}
    f_{tv}(\mathbf{I})=
    \mathbb{E}_{\mathbf{I}} \left[\sum_{i,j}^{H-1,W-1}[(\mathbf{I}_{i+1, j} - \mathbf{I}_{i,j})^2 + (\mathbf{I}_{i, j+1} - \mathbf{I}_{i,j})^2]\right] 
\end{dmath}
where $\mathbf{I}_{i,j}$ is the entry $i,j$ of the image matrix $\mathbf{I}$.

\medskip
\noindent \textbf{Total loss}. The final objective function used to optimize $G$ and $D$ is formulated as follows:
\begin{dmath}
\mathcal{L}_{D} = \lambda_{D}\mathcal{L}_{GAN}  + \lambda_{cls}\mathcal{L}_{cls}
\label{eq:loss_D}
\end{dmath}
\begin{dmath}
\mathcal{L}_{G} = \lambda_{G}\mathcal{L}_{GAN} + \lambda_{rec}\mathcal{L}_{rec} + \lambda_{idt}\mathcal{L}_{idt} + \lambda_{cyc}\mathcal{L}_{cyc} + \lambda_{cls}\mathcal{L}_{\hat{cls}} + \lambda_{tv}\mathcal{L}_{tv}
\label{eq:loss_G}
\end{dmath}
where $\lambda_{D}$, $\lambda_{G}$, $\lambda_{rec}$, $\lambda_{idt}$, $\lambda_{cyc}$, $\lambda_{cls}$, and $\lambda_{tv}$ are hyper-parameters that control the relative importance of each loss term.

\subsection{Rolling Guidance}
\label{Sec:rolling-guidance}
\begin{sloppypar}
While the total variation loss $\mathcal{L}_{tv}$ also enforces the approximated images $\widetilde{\mathbf{I}}_Y$ to be smooth, the source and target images might contain edges and details that have to be preserved. 
Moreover, the $E_1$ and $E_2$ encoders mostly focus on the gesture, failing to learn important details of the context, which might result in blurred images.
Inspired by previous works~\cite{mosinska2018beyond,ma2017pose, zhang2014rolling}, we propose a Rolling Guidance approach to refine the generated image in a two-stage process. 
First, the network generates an initial version $\widehat{\mathbf{I}}_Y$ from input ($\mathbf{I}_X$, $\mathbf{S}_Y$, $C_Y$). 
Then, $\widehat{\mathbf{I}}_Y$ is fed back to $E_2$. Thus, the network generates a refined version of $\widehat{\mathbf{I}}_Y$ from input ($\mathbf{I}_X$, $\mathbf{S}_Y$, $C_Y$, $\widehat{\mathbf{I}}_Y$).
Note that there exists some approaches, like PG$^2$~\cite{ma2017pose}, feed the initial generated image back and concatenated to the source input to learn difference map to refine the results. However, images with a considerable variation in both the foreground and background between source and target images might result in an ill-posed problem. Gesture-to-gesture translation in the wild shows such an issue. For this reason, in \TGAN, we feedback the generated image to $E_2$ to refine the generated image and to learn the condition related features of the target gesture at the same time. This results in better generalization and significantly improves the results as well.
\end{sloppypar}

\section{Experiments}
We compare our model with the state of the art techniques on two hand gesture datasets. 
First, we evaluate our model quantitatively through various widely used metrics that compare the generated image with the ground truth. 
Then, we also evaluate our results qualitatively through a perceptual user study.
We released the resulted dataset and annotations, source code and trained models are available at: \url{https://github.com/yhlleo/TriangleGAN}.

\subsection{Datasets}
The NTU Hand Gesture dataset~\cite{ren2013robust} is a collection of 1,000 RGB-D images recorded with a Kinect sensor. It includes 10 gestures repeated 10 times by 10 subjects under a cluttered background. Image resolution is 640x480. The Creative Senz3D dataset~\cite{memo2018head} is a collection of 1,320 images, where 11 gestures are repeated 30 times by 4 subjects. 
Image resolution is 640x480.

\medskip
\noindent \textbf{Experimental setting.}
We consider two dataset setups in our experimental evaluation.

\medskip
\noindent \textit{Normal} uses the same setup as GestureGAN~\cite{tang2018gesturegan}, in order to directly compare with the state of the art.
GestureGAN authors used only a subset of the datasets (acquired by OpenPose~\cite{cao2018openpose}): 647 images out of 1,000 for NTU Hand Gesture and 494 out of 1,320 for Creative Senz3d. It shows that the state-of-the-art detector OpenPose~\cite{cao2018openpose} fails at detecting key-points for about 50\% of the images.
The resulting number of training and test data pairs obtained are respectively the following:
21,153 and 8,087 for NTU Hand Gesture; 
30,704 and 11,234 for Creative Senz3D.
These numbers are different from those reported in~\cite{tang2018gesturegan} since we here report only the unique pairs without considering flipping and A-to-B reverse ordering.

\medskip
\noindent \textit{Challenging} pushes the limits of our model by ensuring that all the translation pairs "A-to-B" to a specific image "B" are included either in the train or in the test, a condition not ensured by the \emph{normal} setting and, thus, by state of the art.
As a consequence, the model here generates multi-domain images without previous knowledge about it.
We randomly select for the training and test data the following number of pairs: 
22,050 and 13,500 for NTU Hand Gesture;
138,864 and 16,500 for the Creative Senz3D. 



\begin{figure}[ht]
	\centering
	\subfloat{%
		\includegraphics[width=\qwidthSix\columnwidth]{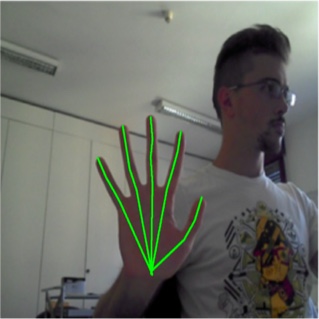}}\hfill
	\subfloat{%
		\includegraphics[width=\qwidthSix\columnwidth]{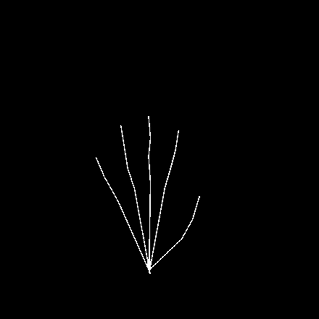}}\hfill
	\subfloat{%
		\includegraphics[width=\qwidthSix\columnwidth]{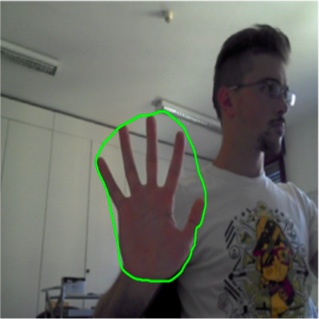}}\hfill
	\subfloat{%
		\includegraphics[width=\qwidthSix\columnwidth]{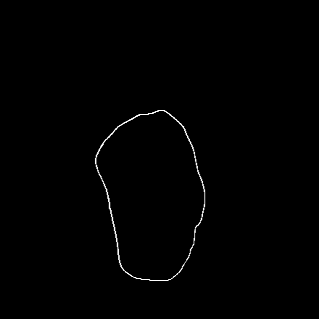}}\hfill
	\subfloat{%
		\includegraphics[width=\qwidthSix\columnwidth]{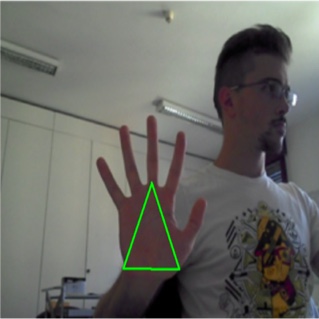}}\hfill
	\subfloat{%
		\includegraphics[width=\qwidthSix\columnwidth]{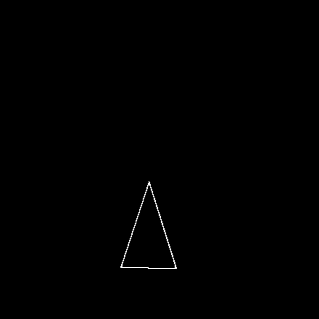}}
	\caption{The three considered shapes of the conditional map, sorted by user drawing effort: from (left) the most difficult to (right) the easiest to draw.}
	\label{fig:maps}
\end{figure}

\medskip
\noindent \textbf{Conditional Maps.}
We consider three possible shapes of the conditional map to prove the effectiveness and generality of our method. Sample images are reported in Figure~\ref{fig:maps}.

\medskip
\noindent \textit{Triangle Map.}
In this type of annotation, the user has to provide an oriented triangle which outlines the size, base and orientation of the hand palm.
This conditional map is easy to draw, as it is possible to provide a simple interface where users can draw a triangle simply specifying the three delimiting points of the triangle, plus its base.
Moreover, the triangle conditional map is category independent, as it does not contain any information about the gesture.
We annotated all images for both datasets with the corresponding triangle conditional maps.

\medskip
\noindent \textit{Boundary Map.}
In the boundary map annotation, the user has to draw the contour of the desired target gesture. This type of annotation is weakly category dependent, since from the conditional map (see Figure~\ref{fig:maps} in the center) it may be possible to infer the target gesture category. However, as this shape is easy to draw, it may be a valid alternative to the skeleton and triangle maps.
We annotated all 1,320 images for the NTU Hand dataset with the corresponding boundary annotations.

\medskip
\noindent \textit{Skeleton Map.}
In the skeleton map, the user is required to draw either a complicated skeleton of the hand gesture or the exact position of the hand gesture key-points.
However, when the target gesture image is available, it is sometimes possible to draw them automatically. 
As in~\cite{tang2018gesturegan}, we obtain the skeleton conditional maps by connecting the key-points obtained through OpenPose, a state of the art hand key-points detector~\cite{simon2017hand, cao2018openpose}.
In the case of \textit{normal} experimental setting, the hand pose is detected for all the 647 and 494 images of the two datasets. Instead, in the case of \textit{challenging} experimental setting, the key-points could not be obtained for over half of the image set. 
To this reason, the skeleton map is considered only in the \textit{normal} experimental setting.
This conditional map is hard to draw, and strongly dependent on the category of hand gesture.

\subsection{Evaluation}
\label{Sec:evaluation}

\textbf{Baseline models.}
As our baseline models we select the state of the art for hand gesture translation \emph{in the wild} GestureGAN~\cite{tang2018gesturegan}. 
Also, we adopt StarGAN~\cite{choi2018stargan}, GANimation~\cite{pumarola2018ganimation}, and PG$^2$~\cite{ma2017pose} as they showed impressive results on multi-domain image-to-image translation. 
Both StarGAN and GANimation learn to use attribute vectors to transfer facial images from one expression to another one. GestureGAN learns to transfer hand gestures via category-dependent skeleton maps.

\medskip
\noindent \textbf{Evaluation metrics.}
We quantitatively evaluate our method performance using two metrics that measure the quality of generated images, namely Peak Signal-to-Noise Ratio (PSNR) and Fréchet Inception Distance (FID)~\cite{NIPS2017_7240}, and the F1-score, which measures whether the generated images depict a consistent category label. 
Moreover, to be comparable with GestureGAN~\cite{tang2018gesturegan}, we employ the Mean Squared Error (MSE) between pixels and the Inception Score (IS)~\cite{gulrajani2017improved}.
However, these two metrics were indicated as highly unstable~\cite{barratt2018note,borji2019pros} 
and they are not strictly related to the quality of generated images. To this reason, we report their results on a separate table and do not further discuss them. 

\medskip
\noindent \textit{PSNR.} It compares two images through their MSE and the maximal pixel intensity ($MAX_I = 255$). It is defined as: $PSNR = 20 \log_{10}(\frac{MAX_I}{\sqrt{MSE}})$.

\medskip
\noindent \textit{FID.} It is defined as the distance between two Gaussian with mean and covariance $(\mu_x, \Sigma_x)$ and $(\mu_y, \Sigma_y)$: $FID(x,y) = || \mu_x- \mu_y ||^2_2 + Tr(\Sigma_x + \Sigma_y - 2(\Sigma_x\Sigma_y)^{1/2})$, where the two Gaussians are defined in the feature space of the Inception model.

\medskip
\noindent \textit{F1.} The F1-score for binary classifiers is defined as $F_1 = (2 p r)/(p+r)$ where $p$ and $r$ are the precision and recall. For multi-class classifiers it can be defined as the sum of 
the F1-scores of each class, weighted by the percentage of labels belonging to the class.
The resulting measure ranges from 0 to 1, where 1 means that all the classes are correctly classified and 0 the opposite. Thus, we use the F1-score to evaluate the category consistence of our generated images. To compute the F1-score, we train a network to recognize hand gestures. Further details are provided in the Implementation Details.

\medskip
\noindent \textit{Perceptual User study.} 
We run a ``fake'' vs ``real'' perceptual user study following the same protocol as~\cite{tang2018gesturegan,yang2018crossing}. Users are shown a pair of images, the original and the generated image, and they are asked to select the fake one. The images are displayed for only 1 second before the user can provide his or her choice. The image pairs are randomly shuffled in order to avoid introducing bias in the results. 
Overall, we collected perceptual annotations from 12 users and each user was asked to vote for 98 image comparisons. Specifically, 12 image translation pairs were selected for each dataset and experimental setting.


\begin{table}[t]
	\centering
	\caption{Quantitative comparison for the gesture-to-gesture translation task in \textit{normal} experimental setting, using the same evaluation metrics as GestureGAN~\cite{tang2018gesturegan}.}
	\resizebox{0.99\linewidth}{!}{ 
	\begin{tabular}{@{}l rrr rrr@{}}
	\toprule
		\textbf{Model}  & 
		\multicolumn{3}{c}{\textbf{NTU Hand Gesture~\cite{ren2013robust}}} & \multicolumn{3}{c}{\textbf{Creative Senz3D~\cite{memo2018head}}} \\
		\cmidrule(r{4pt}){2-4} \cmidrule(l{4pt}){5-7}
		& MSE & PSNR & IS & MSE & PSNR & IS
		\\
		\midrule
		PG$^2$~\cite{ma2017pose}  & 116.10 & 28.24 & 2.42 & 199.44 & 26.51 & 3.37
		\\
		Yan \emph{et al.}~\cite{yan2017skeleton}  & 118.12 & 28.02 & 2.49 & 175.86 & 26.95 & 3.33
		\\
		Ma \emph{et al.}~\cite{ma2018disentangled}  & 113.78 & 30.65 & 2.45 & 183.65 & 26.95 & 3.38
		\\
		PoseGAN \emph{et al.}~\cite{siarohin2018deformable}  & 113.65 & 29.55 & 2.40 & 176.35 & 27.30 & 3.21
		\\
		GestureGAN~\cite{tang2018gesturegan}  &  105.73 & 32.61 & \textbf{2.55} & 169.92 & 27.97 & \textbf{3.41}
		\\
		\TGAN~- no rolling &
		31.80 & 33.57 & 1.92 &
		46.79 & 31.98 & 2.31
		\\
		\TGAN &
		\textbf{15.76} & \textbf{36.51} & 2.00 &
		\textbf{21.73} & \textbf{35.39} & 2.34
		\\
		\bottomrule
	\end{tabular}
	}
	\label{tab:QuantitativeResults_normal}
\end{table}



\begin{table*}[t]
	\centering
	\caption{Quantitative results for the gesture-to-gesture translation task for the two experimental settings.} 
	\begin{tabularx}{\textwidth}{@{}Xccrrrrrrrrrrrrrrrrrrr@{}}
	\toprule
		\textbf{Model}  & \textbf{Experimental setting} & \textbf{Easy to draw map} & \phantom{a} &
		\multicolumn{3}{c}{\textbf{NTU Hand Gesture~\cite{ren2013robust}}} & \phantom{a} & \multicolumn{3}{c}{\textbf{Creative Senz3D~\cite{memo2018head}}} \\
		\cmidrule{5-7} \cmidrule{9-11}
		& & && PSNR & FID & F1 && PSNR & FID & F1
		\\
		\midrule
		GANimation~\cite{pumarola2018ganimation}  &  \multirow{6}{*}{\textit{Normal}}  & \multirow{6}{*}{$\times$} &&  10.03 & 440.45 & 0.08 && 11.11 & 402.99 & 0.23 
		\\
		StarGAN~\cite{choi2018stargan}  &&&&  17.79  & 98.84 & 0.09 && 11.44 & 137.88 & 0.07 
		\\
		PG$^2$~\cite{siarohin2018deformable}  &&&& 21.71 & 66.83 & 0.15 && 21.78 & 122.08 & 0.68
		\\
		GestureGAN~\cite{tang2018gesturegan}  &&&&  34.24 & 15.06 & \textbf{0.93} &  & 28.65 &  54.00 & 0.96
		\\
		\TGAN\ - no rolling &&&&
		34.07 & 20.29 & 0.90 &&
		31.98 & 58.28 & 0.88
		\\
		\TGAN &&&&
		\textbf{36.51} & \textbf{13.73} & \textbf{0.93} &&
		\textbf{35.39} & \textbf{32.04} & \textbf{0.99}
		\\
		\midrule
		\midrule
		PG$^2$~\cite{ma2017pose}  & \multirow{5}{*}{\textit{Challenging}}  & \multirow{5}{*}{\checkmark} && 21.94 & 135.32 & 0.10 && 18.30 & 265.73 & 0.10
		\\
		GestureGAN~\cite{tang2018gesturegan}  &&  && 25.60 & 94.10 & 0.38 & & 18.69 & 254.37 & 0.34
		\\
		GestureGAN$^\dag$  &&  && 27.32 & 59.79 & 0.43 & & 22.24 & 192.77 & 0.38
		\\
		\TGAN\ - no rolling     &&& &    27.14 & 61.14 & 0.43  && 22.77 & 134.54 & 0.33
		\\
		\TGAN     &&&&    \textbf{28.11} & \textbf{47.88} & \textbf{0.61} && \textbf{23.11} & \textbf{101.22} & \textbf{0.58}
		\\
		\bottomrule
	\end{tabularx}
	\label{tab:QuantitativeResults_both}
\end{table*}
\begin{sloppypar}

\medskip
\noindent \textbf{Implementation Details.}
Inspired by previous methods~\cite{zhu2017unpaired,choi2018stargan}, both $E_1$ and $E_2$ are composed of two convolutional layers with the stride size of two for downsampling, $F_{res}$ refers to six residual blocks~\cite{he2016deep}, and $F_{dec}$ is composed of two transposed convolutional layers with the stride size of two for upsampling. 
We train our model using Adam~\cite{kingma2014adam} with $\beta_1=0.5$ and $\beta_2=0.999$ and batch size 4. We use an $n$-dimensional one-hot vector to represent the category label ($n=10$ for NTU dataset and $n=11$ for Senz3D Dataset). For data augmentation we flip the images horizontally with a probability of 0.5 and we reverse the ``A-to-B" direction with a probability of 0.5. The initial learning rate is set to $0.0002$. We train for 20 epochs and linearly decay the rate to zero over the last 10 epochs. To reduce model oscillation~\cite{goodfellow2016nips}, we follow previous works~\cite{shrivastava2017learning,zhu2017unpaired} and update the discriminators using a history of generated images rather than the ones produced by the latest generators. 
We use instance normalization~\cite{ulyanov2016instance} for the generator $G$.
For all the experiments, the weight coefficients for the loss term in Eq.~\ref{eq:loss_D} and Eq.~\ref{eq:loss_G} are set to $\lambda_{D} = 1$, $\lambda_{G} = 2$, 
$\lambda_{cls} = 1$, $\lambda_{rec}=100$, $\lambda_{idt}=10$, $\lambda_{cyc}=10$ and $\lambda_{tv}=1e-5$.
Baseline models are optimized using the same settings described in the respective articles.
We used the source code released by the authors for all competing works, except for GestureGAN, which was implemented from scratch following the description of the original article~\cite{tang2018gesturegan}.
\TGAN~is implemented using the deep learning framework PyTorch. 

To compute the F1-score, we train a network on hand gesture recognition using Inception v3~\cite{szegedy2016rethinking} network fine tuned on the NTU Hand Gesture and Creative Senz3D datasets. 
The network achieves F1-score 0.93 and 0.99 on Creative Senz3D and NTU Hand Gesture test sets, respectively. Additional details on the training can be found in the supplementary materials.

\end{sloppypar}




\begin{figure*}[ht]
	\centering
	\subfloat[Source]{%
		\includegraphics[width=\qwidth\textwidth]{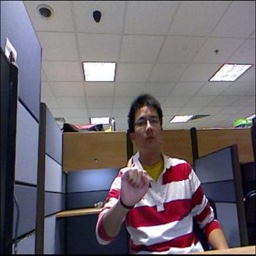}}\hfill
	\subfloat[Target skeleton]{%
		\includegraphics[width=\qwidth\textwidth]{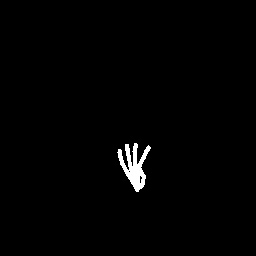}}\hfill
	\subfloat[Ground truth]{%
		\includegraphics[width=\qwidth\textwidth]{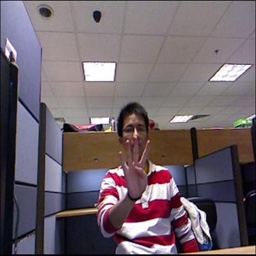}}\hfill
	\subfloat[GANimation~\cite{pumarola2018ganimation}]{%
		\includegraphics[width=\qwidth\textwidth]{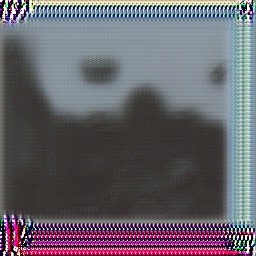}}\hfill
	\subfloat[StarGAN~\cite{choi2018stargan}]{%
		\includegraphics[width=\qwidth\textwidth]{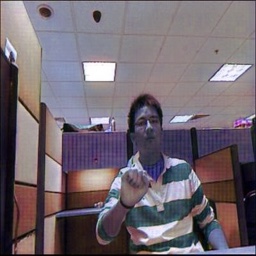}}\hfill
	\subfloat[PG$^2$~\cite{ma2017pose}]{%
		\includegraphics[width=\qwidth\textwidth]{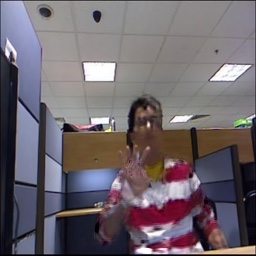}}\hfill
	\subfloat[GestureGAN~\cite{tang2018gesturegan}]{%
		\includegraphics[width=\qwidth\textwidth]{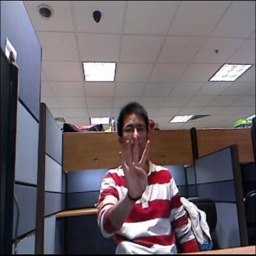}}\hfill
	\subfloat[\boldTGAN ]{%
		\includegraphics[width=\qwidth\textwidth]{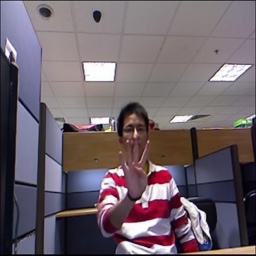}}
	%
    \\
	\subfloat{%
		\includegraphics[width=\qwidth\textwidth]{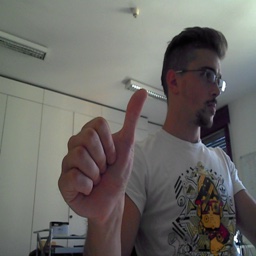}
		}\hfill
	\subfloat{%
		\includegraphics[width=\qwidth\textwidth]{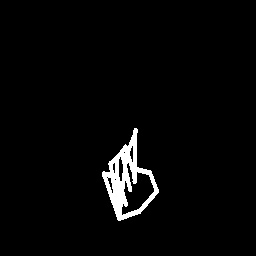}}\hfill
	\subfloat{%
	\includegraphics[width=\qwidth\textwidth]{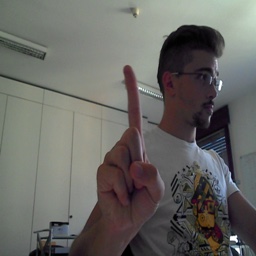}}\hfill
	\subfloat{%
		\includegraphics[width=\qwidth\textwidth]{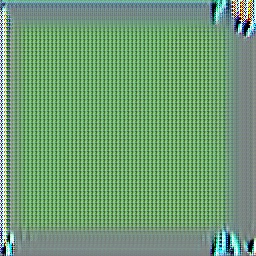}}\hfill
	\subfloat{%
		\includegraphics[width=\qwidth\textwidth]{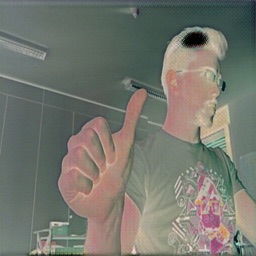}}\hfill
	\subfloat{%
		\includegraphics[width=\qwidth\textwidth]{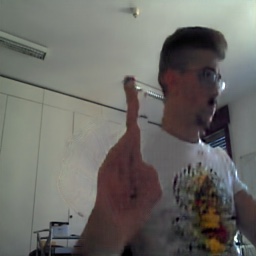}}\hfill
	\subfloat{%
		\includegraphics[width=\qwidth\textwidth]{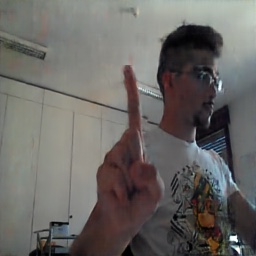}}\hfill
	\subfloat{ 
		\includegraphics[width=\qwidth\textwidth]{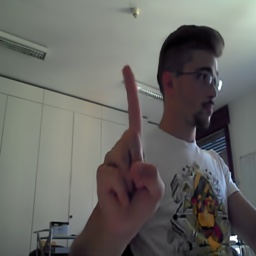}}
	\\
	\subfloat{
		\includegraphics[width=\qwidth\textwidth]{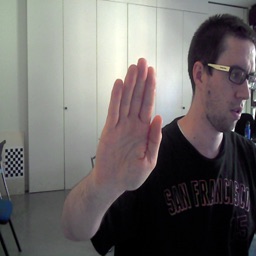}}\hfill
	\subfloat{%
		\includegraphics[width=\qwidth\textwidth]{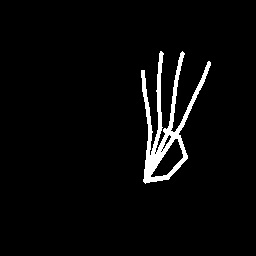}}\hfill
	\subfloat{%
		\includegraphics[width=\qwidth\textwidth]{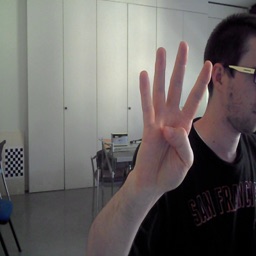}}\hfill
	\subfloat{%
		\includegraphics[width=\qwidth\textwidth]{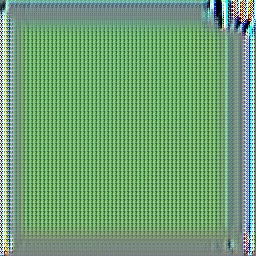}}\hfill
	\subfloat{%
		\includegraphics[width=\qwidth\textwidth]{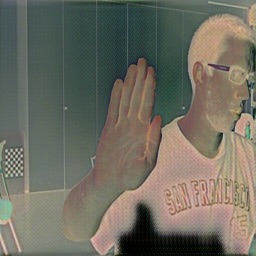}}\hfill
	\subfloat{%
		\includegraphics[width=\qwidth\textwidth]{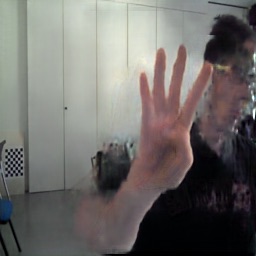}}\hfill
	\subfloat{%
		\includegraphics[width=\qwidth\textwidth]{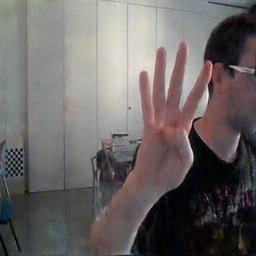}}\hfill
	\subfloat{%
		\includegraphics[width=\qwidth\textwidth]{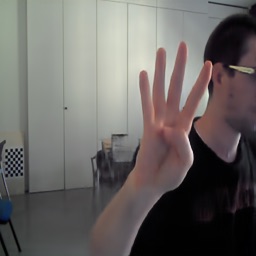}}
	\caption{Qualitative comparison between \TGAN~and competing works in the \textit{normal} experimental setting. NTU Hand Gesture dataset (top two rows) and Creative Senz3D (bottom two rows).}
	\label{fig:res_qual_normal}
\end{figure*}

\begin{figure}[t]
	\centering
	\subfloat[Source]{%
		\includegraphics[width=\qwidthA\columnwidth]{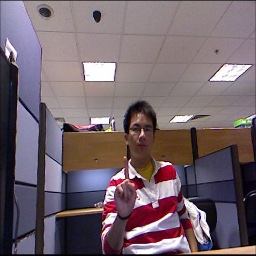}}\hfill
	\subfloat[Target]{%
		\includegraphics[width=\qwidthA\columnwidth]{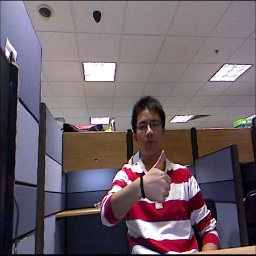}}\hfill
	\subfloat[PG$^2$]{%
		\includegraphics[width=\qwidthA\columnwidth]{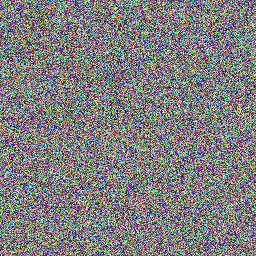}}\hfill
	\subfloat[GANimation]{%
		\includegraphics[width=\qwidthA\columnwidth]{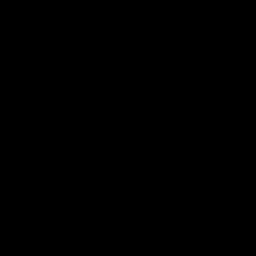}}\hfill
	\subfloat[\boldTGAN]{%
		\includegraphics[width=\qwidthA\columnwidth]{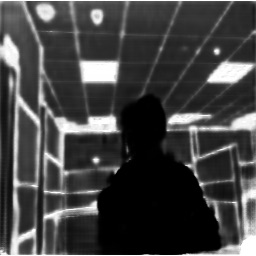}}
	\\
	%
	%
	\subfloat{%
		\includegraphics[width=\qwidthA\columnwidth]{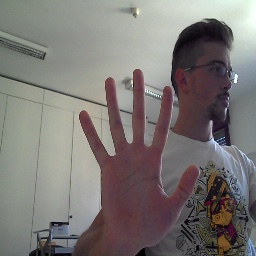}}\hfill
	\subfloat{%
		\includegraphics[width=\qwidthA\columnwidth]{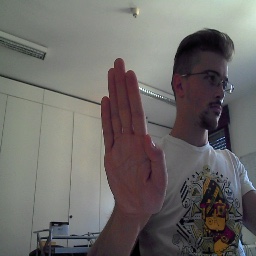}}\hfill
	\subfloat{%
		\includegraphics[width=\qwidthA\columnwidth]{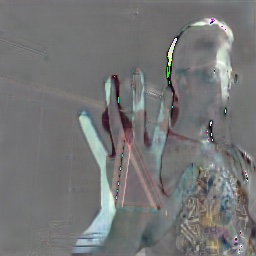}}\hfill
	\subfloat{%
		\includegraphics[width=\qwidthA\columnwidth]{figures/qualitative/S1-G1-16-color-AB-S1-G4-26-color_fake_B_mask.png}}\hfill
	\subfloat{%
		\includegraphics[width=\qwidthA\columnwidth]{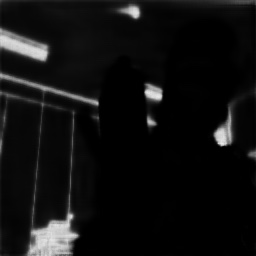}}
	%
	\caption{Masks computed by the various state of the art methods. Specifically, PG$^2$ computes a difference map that is noisy, GANimation fails at computing the attention mask, while \boldTGAN~computes the attention of the pixels that stay constant from source to target images.}
	\label{fig:attention}
\end{figure}


\begin{figure*}[ht]
	\centering
	\begin{minipage}{.88\textwidth}
	\centering
	\subfloat[Source]{%
		\includegraphics[width=\qwidthL\textwidth]{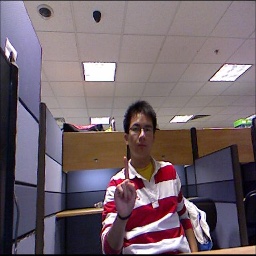}}\hfill
	\subfloat[Target triangle]{%
		\includegraphics[width=\qwidthL\textwidth]{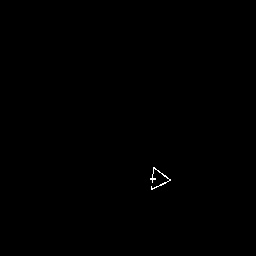}}\hfill
	\subfloat[Category label]{%
		\squarecat{"9"}}\hfill
	\subfloat[Ground truth]{%
		\includegraphics[width=\qwidthL\textwidth]{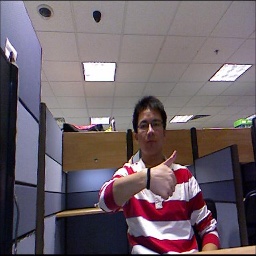}}\hfill
	\subfloat[PG$^2$]{%
		\includegraphics[width=\qwidthL\textwidth]{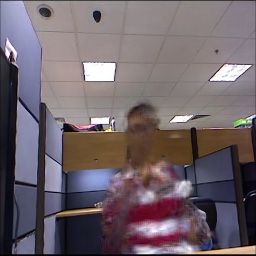}}\hfill
	\subfloat[GestureGAN]{%
		\includegraphics[width=\qwidthL\textwidth]{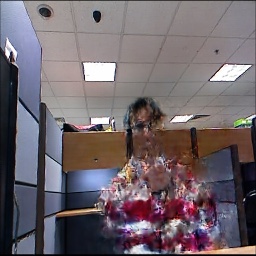}}\hfill
	\subfloat[\boldTGAN]{%
		\includegraphics[width=\qwidthL\textwidth]{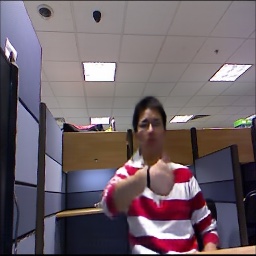}} 
		\\
	\subfloat{%
		\includegraphics[width=\qwidthL\textwidth]{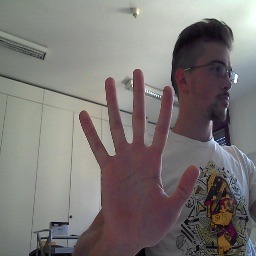}}\hfill
	\subfloat{%
		\includegraphics[width=\qwidthL\textwidth]{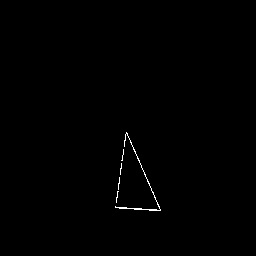}}\hfill
	\subfloat{%
		\squarecat{"4"}}\hfill
	\subfloat{%
		\includegraphics[width=\qwidthL\textwidth]{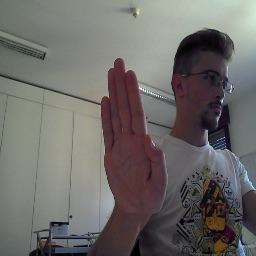}}\hfill
	\subfloat{%
		\includegraphics[width=\qwidthL\textwidth]{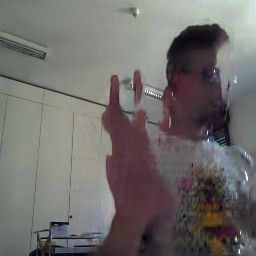}}\hfill
	\subfloat{%
		\includegraphics[width=\qwidthL\textwidth]{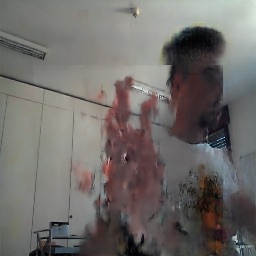}}\hfill
	\subfloat{%
		\includegraphics[width=\qwidthL\textwidth]{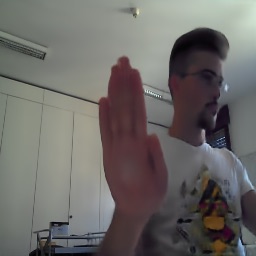}}
	\end{minipage}
	\caption{Qualitative comparison between \TGAN~and competing works in the \textit{challenging} experimental setting. NTU Hand Gesture dataset (top row) and Creative Senz3D (bottom row). 
	}
	\label{fig:res_qual_challenging}
	
\end{figure*}



\section{Results}
\noindent \textbf{Quantitative Results.} 
We begin by directly comparing \TGAN\ with the same experimental protocol and metrics used by our most similar competitor, GestureGAN~\cite{tang2018gesturegan}.
Table~\ref{tab:QuantitativeResults_normal} shows that \TGAN\ performs better than all competing works in terms of MSE and PSNR, especially when the rolling guidance is employed. In terms of IS GestureGAN performs better.
However, the MSE and the IS are not directly related to the quality of the generated images. The MSE is indeed a metric of pixel difference, while the low reliability of the Inception Score is well known~\cite{barratt2018note}.

For this reason, we compare our method with competing works using the PSNR, F1-score and the FID score, both in \textit{normal} and \textit{challenging} experimental settings. These metrics compare the diversity, quality and hand-gesture consistency of the generated images.
Table~\ref{tab:QuantitativeResults_both} shows that \TGAN\ outperforms all competing works, in both the experimental settings and for all the metrics.
In the \emph{normal} setting, compared to GestureGAN, we achieve higher PSNR (36.51 vs 34.24 and 35.39 vs 28.65), F1-score (0.99 vs 0.96) and lower FID (13.73 vs 15.06 and 32.04 vs 54.00) in both datasets. 
Other methods perform particularly poor in terms of F1-score. 
For example, GANimation and StarGAN perform near randomness in the F1-score (random baseline $\sim 0.09$), while \TGAN~ achieves near-perfect performance in Creative Senz3D ($0.99$)  and NTU Hand Gesture ($0.93$).
These outcomes might be related to the fact that GANimation, StarGAN and PG$^2$ are not designed for image translation \emph{in the wild}.

\TGAN\ significantly outperforms the competing methods in the \emph{challenging} setting, where we enforce a stricter division of training and test gestures, and we use the 100\% of the data, differently from GestureGAN's settings.
In particular, the FID score reduces by 49\% and 60\% from GestureGAN in NTU Hand Gesture and Creative Senz3D, respectively. 
In terms of F1-score, the result improves by 61\% and 71\% in NTU Hand Gesture and Creative Senz3D, respectively. 
The PSNR improves by 10\% and 24\% in NTU Hand Gesture and Creative Senz3D, respectively. We note that the rolling guidance applied to GestureGAN (denoted in Table~\ref{tab:QuantitativeResults_both} as GestureGAN$^\dag$) improves the original FID results by 36\% and 24\% in NTU Hand Gesture and Creative Senz3D, respectively. 


Altogether, the quantitative results show that \TGAN\ outperforms the state of the art, both in the \emph{normal} and \emph{challenging} setting. 

\begin{sloppypar}
\medskip
\noindent \textbf{Qualitative Results.} Figure~\ref{fig:res_qual_normal} shows some randomly selected gesture-to-gesture translations in the \textit{normal} experimental setting. 
Both GestureGAN~\cite{tang2018gesturegan} and \TGAN\ produce sharper output results while the output gestures from PG$^2$~\cite{ma2017pose} are very blurry. \TGAN\ also produces a better define sharper background than GestureGAN. 
StarGAN and GANimation, however, fail to produce gestures from the provided conditional map.

We further inspect the reason behind the poor results of PG$^2$ and GANimation. 
These methods focus on multi-domain translation and are specifically tailored to the task of facial expression translation and to the cases where the input and target objects are aligned. 
Figure~\ref{fig:attention} depicts two sample cases of the difference and attention masks generated by these methods. It can be seen that PG$^2$ fails at finding the difference mask, especially in the NTU Hand Gesture dataset (top row). Similarly, GANimation generates an empty attention mask, which might also be the cause of its poor results in both Figure~\ref{fig:res_qual_normal} and Table~\ref{tab:QuantitativeResults_both}. 
\TGAN, instead, learns to focus on the differences between the source and target images.

Figure~\ref{fig:res_qual_challenging} shows the results in the \textit{challenging} setting. 
\TGAN\ generates sharper images with recognizable hand gestures, while other methods such as GestureGAN fails at it. 
This result is in line with the F1-scores reported in Table~\ref{tab:QuantitativeResults_both} (bottom), which range within 0.10 and 0.38 for the competing works, and within 0.58 and 0.61 in case of \TGAN. 
Both qualitative and quantitative results confirm that state-of-the-art methods are not adapted to perform gesture translation in the challenging settings, i.e. where a user-friendly category independent conditional map is provided, and where the network is asked to translate to an unseen gesture for a given user.
We refer to the Supplementary Material for additional qualitative Figures.
\end{sloppypar}


\begin{figure}[!htp]
	\centering
	\subfloat[Source]{%
		\includegraphics[width=\qwidthA\columnwidth]{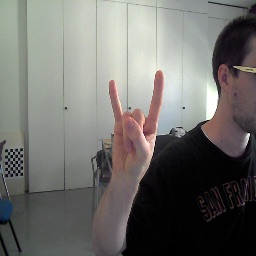}}\hfill
	\subfloat[Target map]{%
		\includegraphics[width=\qwidthA\columnwidth]{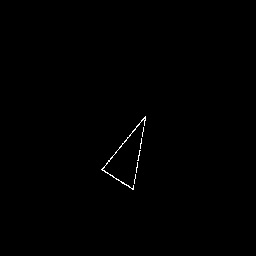}}\hfill
	\subfloat["7"]{%
		\includegraphics[width=\qwidthA\columnwidth]{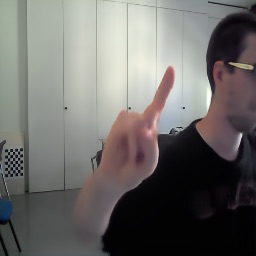}}\hfill
	\subfloat["9"]{%
		\includegraphics[width=\qwidthA\columnwidth]{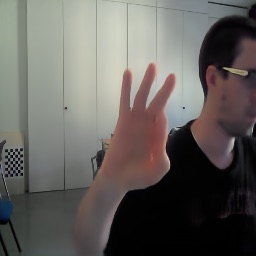}}\hfill
	\subfloat["11"]{%
		\includegraphics[width=\qwidthA\columnwidth]{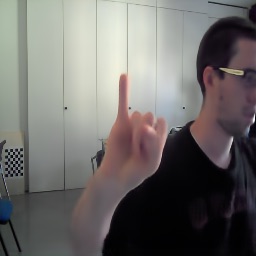}}
	\\ 
	\subfloat{%
		\includegraphics[width=\qwidthA\columnwidth]{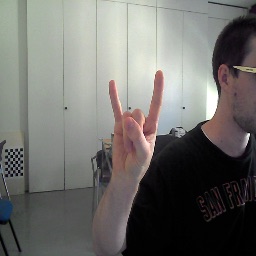}}\hfill
	\subfloat{%
		\includegraphics[width=\qwidthA\columnwidth]{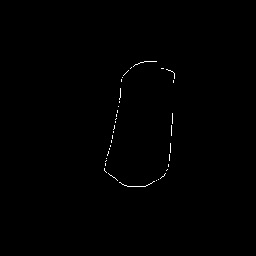}}\hfill
	\subfloat{%
		\includegraphics[width=\qwidthA\columnwidth]{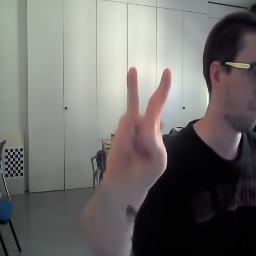}}\hfill
	\subfloat{%
		\includegraphics[width=\qwidthA\columnwidth]{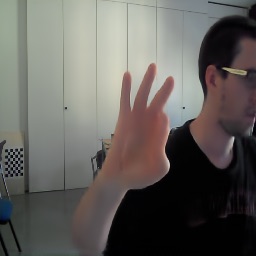}}\hfill
	\subfloat{%
		\includegraphics[width=\qwidthA\columnwidth]{figures/qualitative/diversity/BOS2-G6-4-color-S2-G9-5-color-7_fake_B2_masked.jpeg}}
	\\ 
	\subfloat{%
		\includegraphics[width=\qwidthA\columnwidth]{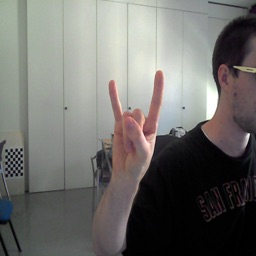}}\hfill
	\subfloat{%
		\includegraphics[width=\qwidthA\columnwidth]{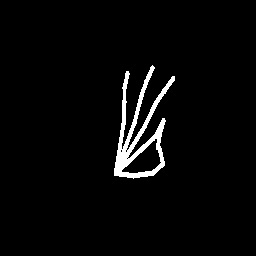}}\hfill
	\subfloat{%
		\includegraphics[width=\qwidthA\columnwidth]{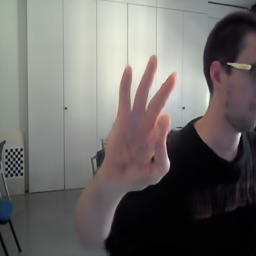}}\hfill
	\subfloat{%
		\includegraphics[width=\qwidthA\columnwidth]{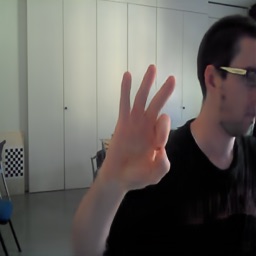}}\hfill
	\subfloat{%
		\includegraphics[width=\qwidthA\columnwidth]{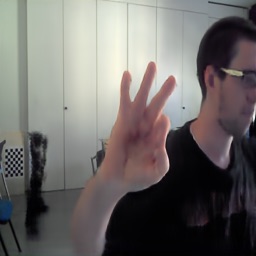}}
	\caption{\TGAN~decouples the conditional map from the category label of the hand gesture. The same conditional map can be used with different category labels to generate multiple images.  
	}
	\label{fig:diversity}

\end{figure}
\begin{figure}[!ht]
	\centering
	\renewcommand{\tabcolsep}{1pt}
	\subfloat{\footnotesize\begin{tabular}{>{\centering\arraybackslash}p{0.39\columnwidth}>{\centering\arraybackslash}p{0.58\columnwidth}}
         \textbf{Source} & \textbf{Conditional maps and generated images} \\ 
    \end{tabular}}\\
	%
    \subfloat{%
		\includegraphics[width=0.39\columnwidth,height=0.396\columnwidth]{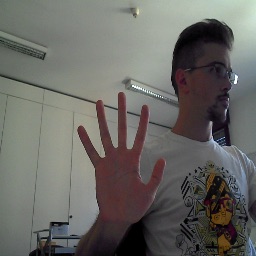}}\hfill
	\subfloat{\begin{minipage}{\qwidthA\columnwidth}%
			\setlength{\lineskip}{3pt}%
			\includegraphics[width=\columnwidth]{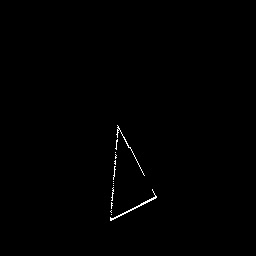}\\
			\includegraphics[width=\columnwidth]{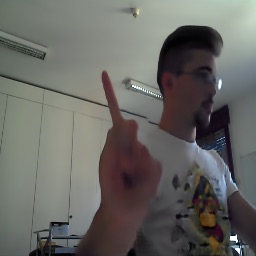}
	\end{minipage}}\hfill
	\subfloat{\begin{minipage}{\qwidthA\columnwidth}%
			\setlength{\lineskip}{3pt}%
			\includegraphics[width=\columnwidth]{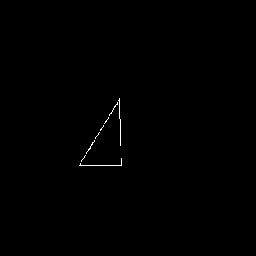}\\
			\includegraphics[width=\columnwidth]{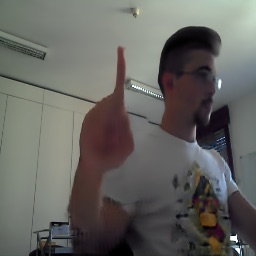}
	\end{minipage}}\hfill
	\subfloat{\begin{minipage}{\qwidthA\columnwidth}%
			\setlength{\lineskip}{3pt}%
			\includegraphics[width=\columnwidth]{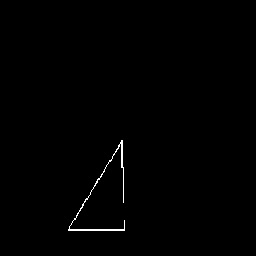}\\
			\includegraphics[width=\columnwidth]{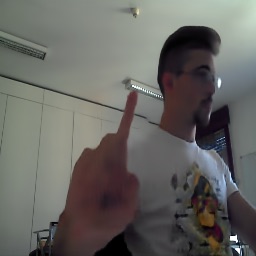}
	\end{minipage}}
	\caption{
	\TGAN~generates images with the same conditional map that is rotated, shifted and resized. 
	}
	\label{fig:diversity2}
\end{figure}




\medskip
\noindent \textbf{Diversity Study.} \TGAN\ decouples the desired hand gesture category, specified through a class number, from its location, scale and orientation, which are specified by a conditional map.
This means that users can use the same conditional map with different hand gesture categories to generate multiple images.
Figure~\ref{fig:diversity} shows that \TGAN\ can generate three different distinguishable hand gestures by using the same \emph{triangle} conditional map and different category numbers.
Instead, when using non category-independent shapes are used (e.g. boundary or skeleton), \TGAN\ fails to synthesize several hand gestures categories from the same conditional map. 
As mentioned before, \emph{Boundary} maps are weakly category dependent, as their shape might suggest the type of gesture, while \emph{Skeleton} and \emph{Key-points} maps are category dependent. 

Furthermore, we test the performance of \TGAN\ with the same source image and category, but with different conditional maps. 
For this test, we manually draw three triangle conditional maps with arbitrary size, location and orientation.
Figure~\ref{fig:diversity2} shows that our method faithfully synthesizes the target gestures in all cases.
Altogether, we show that users can generate hand gestures \emph{in the wild} with much less effort than state-of-the-art models, that require complex annotations that are dependent on the specific hand gesture users want to generate.

\begin{table}[t]
    \caption{Perceptual user study. Percentage of times, on average, when the translated images are selected as ``real'' by users, in the ``fake'' vs. ``real'' comparison.}
    \centering
    \small
    \begin{tabularx}{\columnwidth}{@{}X rr rrr@{}}
    \toprule 
        \textbf{Model} & \multicolumn{2}{c}{\textbf{NTU Hand Gesture~\cite{ren2013robust}}} & 
        \multicolumn{2}{c}{\textbf{Creative Senz3D~\cite{memo2018head}}}  \\
        \cmidrule(r{4pt}){2-3} \cmidrule(l{4pt}){4-5}
        & \textit{Normal} & \textit{Challenging} & \textit{Normal} & \textit{Challenging} \\
         \midrule
          GestureGAN & 36.54\% & 3.21\% & 3.85\% & 0.64\% \\
          \TGAN & \textbf{44.32\%} & \textbf{7.05\%} & \textbf{16.03\%} & \textbf{3.85\%} \\
          \bottomrule
    \end{tabularx}
    \label{tab:userstudy}
\end{table}

\medskip
\noindent \textbf{Perception User Study.}
Table~\ref{tab:userstudy} shows the outcome of the perceptual user study, where we report the percentage of times when the translated image wins against the original target image in the real vs fake comparison, i.e when the original image is selected as fake. It can be seen that \TGAN\ outperforms GestureGAN~\cite{tang2018gesturegan} in both experimental settings. 
This means that \TGAN\ generates higher quality images that can be mistaken with the real ones at higher rates than competitors.

\begin{table}[t]
	\centering
	\small
	\caption{Performance degradation of \boldTGAN\ by removing the rolling guidance approach. 
	}
	\begin{tabularx}{\columnwidth}{@{}X c rrr@{}}
	\toprule
		\textbf{Conditional map}   & \textbf{Category} & 
		\multicolumn{3}{c}{\textbf{Creative Senz3D~\cite{memo2018head}}} \\ 
		\cmidrule{3-5} 
		& \textbf{independent} & PSNR & FID & F1 
		\\
		\midrule
		\emph{Triangle} & \checkmark & 1.53\% \textcolor{red}{\textdownarrow}  & 24.77\% \textcolor{red}{\textuparrow} & 75.76\% \textcolor{red}{\textdownarrow}
		\\
		\emph{Boundary} & $\sim$ &
		 1.84\% \textcolor{red}{\textdownarrow}  & 35.37\% \textcolor{red}{\textuparrow} & 58.06\% \textcolor{red}{\textdownarrow}
		\\
		\emph{Skeleton} & $\times$ &
		 10.66\% \textcolor{red}{\textdownarrow}  & 81.90\% \textcolor{red}{\textuparrow} & 12.50\% \textcolor{red}{\textdownarrow}
		\\
		\bottomrule
	\end{tabularx}
	\label{tab:ablation}
	\vspace{-0.3cm}
\end{table}

\medskip
\noindent \textbf{Ablation study.} We further investigate the effect of the rolling guidance approach by removing this component from \TGAN. 
In the quantitative results, Table~\ref{tab:QuantitativeResults_both} it can be seen that the rolling guidance allows to significantly improves the quality of the generated images.
In Table~\ref{tab:ablation} we report the degradation of our method performance without rolling guidance, on Creative Senz3D dataset, for the three types of annotation maps. 
Specifically, we observe 24.77\%, 35.37\% and 81.90\% worse (increased) FID score for the \emph{Triangle}, \emph{Boundary} and \emph{Skeleton} maps, respectively. 
While the F1-score decreases by 75.76\% and 58.06\% on the \emph{Triangle} and \emph{Boundary} maps respectively, for \emph{Skeleton} maps it decreases by 12.50\%. In terms of PSNR the degradation is less significant for the \emph{Triangle} and \emph{Boundary}, but higher (10.66\%) for \emph{Skeleton} maps.
Finally, we observed that a single rolling is sufficient to improve the generated results, while \TGAN\ does not benefit from additional rolling iterations.



\section{Conclusion}
We have presented a novel GAN architecture for gesture-to-gesture translation \textit{in the wild}. Our model decouples the conditional input into a category label and an easy-to-draw conditional map.
The proposed attention module and rolling guidance approach allow generating sharper images \textit{in the wild}, faithfully generating the target gesture while preserving the background context.  
Experiments on two public datasets have shown that \TGAN\ outperforms state of the art both quantitatively and qualitatively.
Moreover, it allows the use of simple-to-draw and category-independent conditional maps, such as triangles.
This significantly reduces the annotation effort both at drawing time and allowing to use a single map for multiple gesture translations. The proposed framework is not limited to category labels but can also be used through embedding, learned from the data, that can express the gesture characteristics. In future work users could easily provide a reference image instead of a label, and translate the original image to the target gesture expressed by the reference image.

Our approach is especially important when the target image, and thus the target conditional map, is not available, which is the typical scenario in photo-editing software. 


%
\begin{acks}
We gratefully acknowledge NVIDIA Corporation for the donation of the Titan X GPUs and Fondazione Caritro for supporting the SMARTourism project.
\end{acks}

%

\bibliographystyle{ACM-Reference-Format}
\balance
\bibliography{arxiv}

\appendix

\newpage
\section{Network Architecture}

The network architectures of \TGAN~are shown in Table~\ref{tab:generator-architecture} and Table~\ref{tab:discriminator-architecture}. For the generator network, we use instance normalization in all layers except the last output layer. For the discriminator network, we use Leaky ReLU~\cite{xu2015empirical} with a negative slope of 0.01. We use the following notations: $n_c$: the number of gesture category, $n_d$: the channels of conditional map ($n_d=1$). N: the number of output channels, K: kernel size, S: stride size, P: padding size, IN: instance normalization. Note that $E_2$ part takes ($\mathbf{S}_Y$, $C_Y$) as input with $n_c$+$n_d$ channels in the first stage, and takes ($\mathbf{S}_Y$, $C_Y$, $\widehat{\mathbf{I}}_Y$) as input with $n_c$+$n_d$+3 channels in the second stage.

\section{WGAN Loss}

During the training, we try the WGAN loss~\cite{arjovsky2017wasserstein} $\mathcal{L}_{WGAN}$ that uses the Earth Mover Distance to minimize the differences of the distributions, and a gradient penalty $\mathcal{L}_{gp}$ that improves the stability with respect to other adversarial losses~\cite{gulrajani2017improved}. The WGAN and Gradient Penalty loss are formalized as follows:
\begin{equation}
\begin{aligned}
    \mathcal{L}_{WGAN} = & \mathbb{E}_{\mathbf{I}_Y, \mathbf{S}_Y\sim \mathbb{P}}[D_{prob}(\mathbf{I}_Y,\mathbf{S}_Y)]- \\ 
    & \mathbb{E}_{\mathbf{I}_X,\mathbf{S}_Y, C_Y\sim \mathbb{P}}[D_{prob}(G(\mathbf{I}_X, \mathbf{S}_Y, C_Y),\mathbf{S}_Y)] \\
\end{aligned}
\end{equation}
\begin{equation}
    \mathcal{L}_{gp} =  \mathbb{E}_{\bar{I} \sim \mathbb{P}_{\bar{I}}, \mathbf{S}_Y\sim \mathbb{P}}
    [(\| \triangledown_{\bar{\mathbf{I}}} D_{prob}(\bar{\mathbf{I}}, \mathbf{S}_Y)\|_2 -1)^2]
\end{equation}
where $\bar{\mathbf{I}}$ is uniformly sampled as a linear combination between the ground truth $\textbf{I}_Y$ and the generated $\hat{\mathbf{I}}_Y$. However, there are no significant improvements in the experiments compared with using $L_{GAN}$. 

\section{F1-score Discriminator details}
To compute the F1-score, we train a network on hand gesture recognition using Inception v3~\cite{szegedy2016rethinking} network fine tuned on the NTU Hand Gesture and Creative Senz3D datasets. We freeze all layers but the last three (the fully connected and the two Inception modules) and fine tune the network using 10\% of the data for validation, 10\% for testing and the rest is used for training. 
For training we used SGD with learning rate 0.001, batch-size 64, and momentum 0.9. We use random rotation (until $40\degree$) and flip as data augmentation with probability 0.5.
Given in input the \TGAN~generated images, this network is expected to predict the ground truth category label of the gesture.

\section{FID score}
We note the previous state of the art GestureGAN~\cite{tang2018gesturegan} reports lower FID scores than what we report in our manuscript. We verified that this is due to a software bug in the FID score implementation in PyTorch (\url{https://github.com/mseitzer/pytorch-fid/}). In particular, this implementation used a wrong input normalization that scaled the input in the range [0,0.5):
\begin{lstlisting}
  x[:,0]=x[:,0]*(0.229/0.5) + (0.485-0.5)/0.5
  x[:,1]=x[:,1]*(0.224/0.5) + (0.456-0.5)/0.5
  x[:,2]=x[:,2]*(0.225/0.5) + (0.406-0.5)/0.5
\end{lstlisting}
This incorrect implementation was later modified in May 27th 2019 into $x = 2 * x - 1$~\cite{mseitzer}. We used this corrected implementation, which results in higher FID scores. Table \ref{tab:QuantitativeResults_wrong} reports both the scores. The column of FID* shows the result of the incorrect implementation.

\begin{table}[!htp]
	\centering
	\caption{Quantitative results for the gesture-to-gesture translation task for the two experimental settings. Here we describe the difference between the FID score and the \texttt{mseitzer}'s incorrect implementation (FID*), which was used in \cite{tang2018gesturegan}. }
	\begin{tabularx}{\columnwidth}{@{}Xrrrrr@{}}
	\toprule
		\textbf{Model}  &
		\multicolumn{2}{c}{\textbf{NTU~\cite{ren2013robust}}} & \phantom{a} & \multicolumn{2}{c}{\textbf{Senz3D~\cite{memo2018head}}} \\
		\cmidrule{2-3} \cmidrule{5-6}
		& FID* & FID && FID* & FID 
		\\
		\midrule
		\emph{Normal setting}\\
		GANimation~\cite{pumarola2018ganimation}  & 403.63 & 440.45 && 333.53 & 402.99 \\
		StarGAN~\cite{choi2018stargan}  & 91.81  & 98.84 &&  105.86 & 137.88\\
		PG$^2$~\cite{siarohin2018deformable}  & 42.98  & 66.83 && 119.00 & 122.08 \\
		GestureGAN~\cite{tang2018gesturegan}  & 6.99 & 15.06 &&   29.75 & 54.00 \\
		\TGAN\ - no rolling & 13.32 & 20.29 && 45.12 & 58.28 \\
		\TGAN & \textbf{6.63} & \textbf{13.73} && \textbf{16.8} & \textbf{32.04}
		\\ \midrule \midrule
		\emph{Challenging setting}\\
		PG$^2$~\cite{ma2017pose}  & 144.18 & 135.32 &&  185.92  & 265.73 \\
		GestureGAN~\cite{tang2018gesturegan}    & 69.29  & 94.10 && 183.02 & 254.37  \\
		GestureGAN$^\dag$   & 45.42 & 59.79 &&  114.85 & 192.77 \\
		\TGAN\ - no rolling   & 55.76  & 61.14 && 122.34 & 134.54 \\
		\TGAN   & \textbf{40.02}  & \textbf{47.88} && \textbf{87.32} &  \textbf{101.22} \\
		\bottomrule
	\end{tabularx}
	\label{tab:QuantitativeResults_wrong}
\end{table}

\begin{table*}[!htp]
    \ra{1.2}
	\centering
	\caption{Generator network architecture.}
	\begin{tabular}{@{}lll@{}}
	    \toprule
	    \textbf{Part} & \textbf{Input} $\rightarrow$ \textbf{Output Shape} & \textbf{Layer Information} \\ \hline
		\multirow{3}{*}{$E_1$} & ($h$, $w$, 3) $\rightarrow$ ($h$, $w$, 64) & CONV-(N64, K7x7, S1, P3), IN, ReLU \\
		& ($h$, $w$, 64) $\rightarrow$ ($\frac{h}{2}$, $\frac{w}{2}$, 128) & CONV-(N128, K3x3, S2, P1), IN, ReLU \\
		& ($\frac{h}{2}$, $\frac{w}{2}$, 128) $\rightarrow$ ($\frac{h}{4}$, $\frac{w}{4}$, 256) & CONV-(N256, K3x3, S2, P1), IN, ReLU \\ \hline
		\multirow{3}{*}{$E_2$} & ($h$, $w$, $n_c$+$n_d$+(3)) $\rightarrow$ ($h$, $w$, 64) & CONV-(N64, K7x7, S1, P3), IN, ReLU \\
		& ($h$, $w$, 64) $\rightarrow$ ($\frac{h}{2}$, $\frac{w}{2}$, 64) & CONV-(N64, K3x3, S2, P1), IN, ReLU \\
		& ($\frac{h}{2}$, $\frac{w}{2}$, 64) $\rightarrow$ ($\frac{h}{4}$, $\frac{w}{4}$, 64) & CONV-(N64, K3x3, S2, P1), IN, ReLU \\ \hline	
		Concat+CONV & ($\frac{h}{2}$, $\frac{w}{2}$, 64+256) $\rightarrow$ ($\frac{h}{4}$, $\frac{w}{4}$, 256) & CONV-(N256, K3x3, S1, P1), IN, ReLU \\ \hline
		\multirow{6}{*}{$F_{res}$} & ($\frac{h}{4}$, $\frac{w}{4}$, 256) $\rightarrow$ ($\frac{h}{4}$, $\frac{w}{4}$, 256) & Residual Block: CONV-(N256, K3x3, S1, P1), IN, ReLU \\
		& ($\frac{h}{4}$, $\frac{w}{4}$, 256) $\rightarrow$ ($\frac{h}{4}$, $\frac{w}{4}$, 256) & Residual Block: CONV-(N256, K3x3, S1, P1), IN, ReLU \\
		& ($\frac{h}{4}$, $\frac{w}{4}$, 256) $\rightarrow$ ($\frac{h}{4}$, $\frac{w}{4}$, 256) & Residual Block: CONV-(N256, K3x3, S1, P1), IN, ReLU \\
		& ($\frac{h}{4}$, $\frac{w}{4}$, 256) $\rightarrow$ ($\frac{h}{4}$, $\frac{w}{4}$, 256) & Residual Block: CONV-(N256, K3x3, S1, P1), IN, ReLU \\
		& ($\frac{h}{4}$, $\frac{w}{4}$, 256) $\rightarrow$ ($\frac{h}{4}$, $\frac{w}{4}$, 256) & Residual Block: CONV-(N256, K3x3, S1, P1), IN, ReLU \\
		& ($\frac{h}{4}$, $\frac{w}{4}$, 256) $\rightarrow$ ($\frac{h}{4}$, $\frac{w}{4}$, 256) & Residual Block: CONV-(N256, K3x3, S1, P1), IN, ReLU \\ \hline
		Concat+CONV & ($\frac{h}{4}$, $\frac{w}{4}$, 64+256) $\rightarrow$ ($\frac{h}{4}$, $\frac{w}{4}$, 256) & CONV-(N256, K3x3, S1, P1), IN, ReLU \\ \hline
		\multirow{2}{*}{$F_{dec}$} & ($\frac{h}{4}$, $\frac{w}{4}$, 256) $\rightarrow$ ($\frac{h}{2}$, $\frac{w}{2}$, 128) & DECONV-(N128, K4x4, S2, P1), IN, ReLU \\ 
		& ($\frac{h}{2}$, $\frac{w}{2}$, 128) $\rightarrow$ ($h$, $w$, 64) & DECONV-(N64, K4x4, S2, P1), IN, ReLU \\ \hline
		$G_C$ & ($h$, $w$, 64) $\rightarrow$ ($h$, $w$, 3) & CONV-(N3, K7x7, S1, P3), Tanh\\
		$G_A$ & ($h$, $w$, 64) $\rightarrow$ ($h$, $w$, 1) & CONV-(N1, K7x7, S1, P3), Sigmoid\\
		\bottomrule
	\end{tabular}
	\label{tab:generator-architecture}
\end{table*}

\begin{table*}[!htp]
    \ra{1.2}
	\centering
	\caption{Discriminator network architecture.}
	\begin{tabular}{@{}lll@{}}
	    \toprule
	    \textbf{Layer} & \textbf{Input} $\rightarrow$ \textbf{Output Shape} & \textbf{Layer Information} \\ \hline
		Input Layer & ($h$, $w$, 3+$n_d$) $\rightarrow$ ($\frac{h}{2}$, $\frac{w}{2}$, 64) & CONV-(N64, K4$\times$4, S2, P1), Leaky ReLU \\ \hline
		Hidden Layer & ($\frac{h}{2}$, $\frac{w}{2}$, 64) $\rightarrow$ ($\frac{h}{4}$, $\frac{w}{4}$, 128) & CONV-(N128, K4$\times$4, S2, P1), Leaky ReLU \\
		Hidden Layer & ($\frac{h}{4}$, $\frac{w}{4}$, 128) $\rightarrow$ ($\frac{h}{8}$, $\frac{w}{8}$, 256) & CONV-(N256, K4$\times$4, S2, P1), Leaky ReLU \\ 
		Hidden Layer & ($\frac{h}{8}$, $\frac{w}{8}$, 256) $\rightarrow$ ($\frac{h}{16}$, $\frac{w}{16}$, 512) & CONV-(N512, K4$\times$4, S2, P1), Leaky ReLU \\ 
		Hidden Layer & ($\frac{h}{16}$, $\frac{w}{16}$, 512) $\rightarrow$ ($\frac{h}{32}$, $\frac{w}{32}$, 1024) & CONV-(N1024, K4$\times$4, S2, P1), Leaky ReLU \\ 
		Hidden Layer & ($\frac{h}{32}$, $\frac{w}{32}$, 1024) $\rightarrow$ ($\frac{h}{64}$, $\frac{w}{64}$, 2048) & CONV-(N2048, K4$\times$4, S2, P1), Leaky ReLU \\ \hline
		$D_{cls}$ & ($\frac{h}{64}$, $\frac{w}{64}$, 2048) $\rightarrow$ (1, 1, $n_c$) & CONV-(N($n_c$), K$\frac{h}{64} \times \frac{w}{64}$, S1, P0)\\
		$D_{prob}$ & ($\frac{h}{64}$, $\frac{w}{64}$, 64) $\rightarrow$ ($\frac{h}{64}$, $\frac{w}{64}$, 1) & CONV-(N1, K4$\times$4, S1, P1)\\
		\bottomrule
	\end{tabular}
	\label{tab:discriminator-architecture}
\end{table*}

\section{Additional Qualitative Results}

\subsection{More Comparisons}

\begin{figure*}[!ht]
	\centering
	\subfloat[Source]{%
		\includegraphics[width=\qwidth\textwidth]{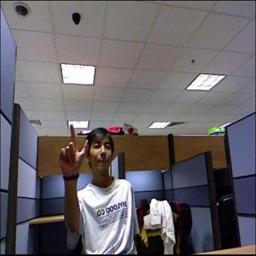}}\hfill
	\subfloat[Target skeleton]{%
		\includegraphics[width=\qwidth\textwidth]{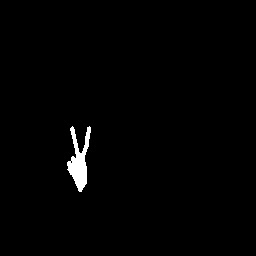}}\hfill
	\subfloat[Ground truth]{%
		\includegraphics[width=\qwidth\textwidth]{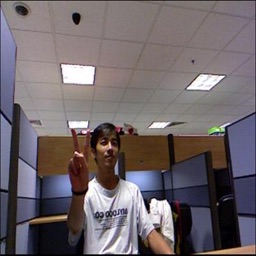}}\hfill
	\subfloat[GANimation~\cite{pumarola2018ganimation}]{%
		\includegraphics[width=\qwidth\textwidth]{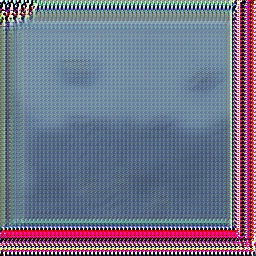}}\hfill
	\subfloat[StarGAN~\cite{choi2018stargan}]{%
		\includegraphics[width=\qwidth\textwidth]{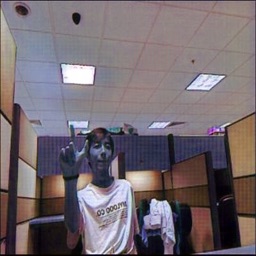}}\hfill
	\subfloat[PG$^2$~\cite{ma2017pose}]{%
		\includegraphics[width=\qwidth\textwidth]{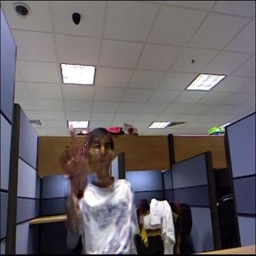}}\hfill
	\subfloat[GestureGAN~\cite{tang2018gesturegan}]{%
		\includegraphics[width=\qwidth\textwidth]{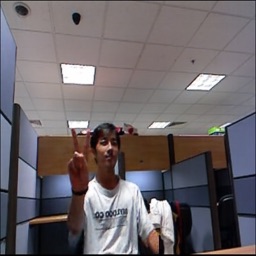}}\hfill
	\subfloat[\boldTGAN ]{%
		\includegraphics[width=\qwidth\textwidth]{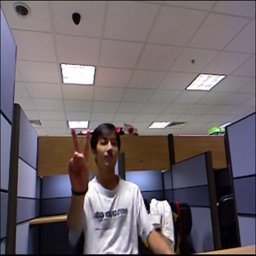}}
	\\
	\subfloat{%
		\includegraphics[width=\qwidth\textwidth]{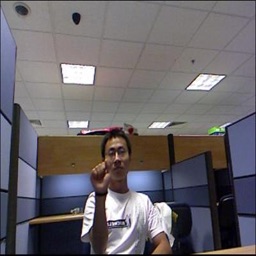}}\hfill
	\subfloat{%
		\includegraphics[width=\qwidth\textwidth]{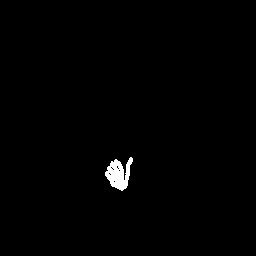}}\hfill
	\subfloat{%
		\includegraphics[width=\qwidth\textwidth]{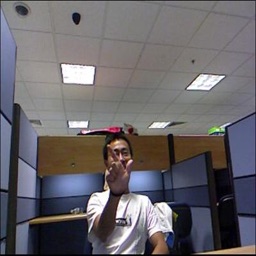}}\hfill
	\subfloat{%
		\includegraphics[width=\qwidth\textwidth]{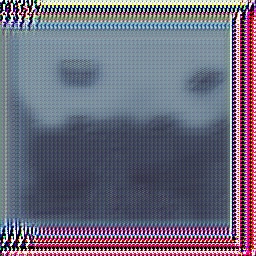}}\hfill
	\subfloat{%
		\includegraphics[width=\qwidth\textwidth]{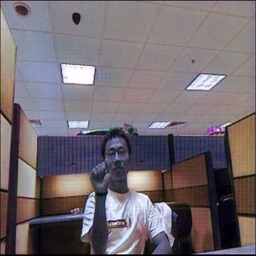}}\hfill
	\subfloat{%
		\includegraphics[width=\qwidth\textwidth]{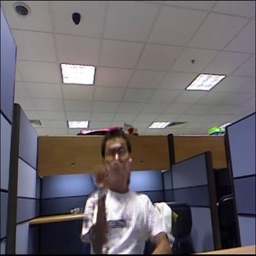}}\hfill
	\subfloat{%
		\includegraphics[width=\qwidth\textwidth]{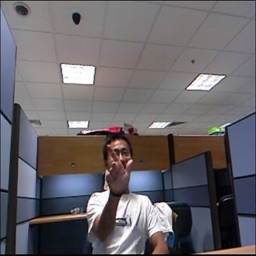}}\hfill
	\subfloat{%
		\includegraphics[width=\qwidth\textwidth]{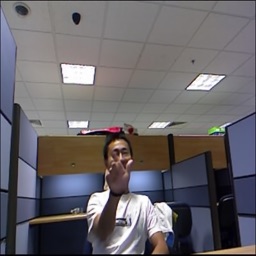}}
	\\
	\subfloat{%
		\includegraphics[width=\qwidth\textwidth]{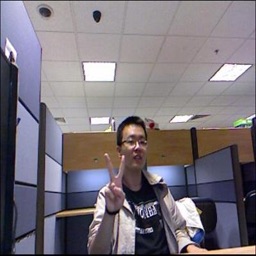}}\hfill
	\subfloat{%
		\includegraphics[width=\qwidth\textwidth]{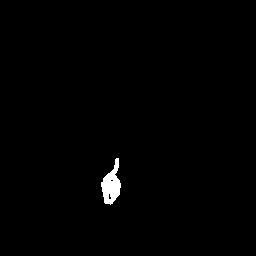}}\hfill
	\subfloat{%
		\includegraphics[width=\qwidth\textwidth]{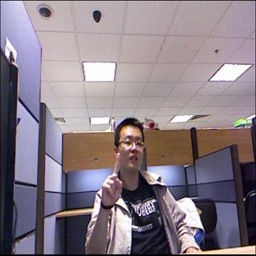}}\hfill
	\subfloat{%
		\includegraphics[width=\qwidth\textwidth]{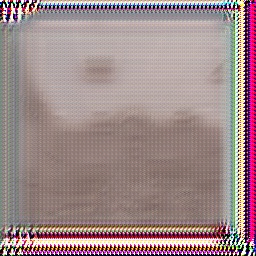}}\hfill
	\subfloat{%
		\includegraphics[width=\qwidth\textwidth]{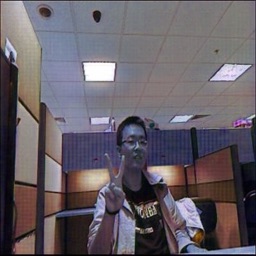}}\hfill
	\subfloat{%
		\includegraphics[width=\qwidth\textwidth]{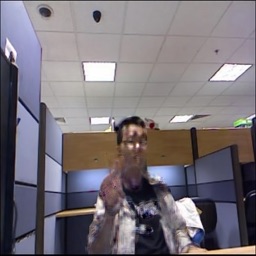}}\hfill
	\subfloat{%
		\includegraphics[width=\qwidth\textwidth]{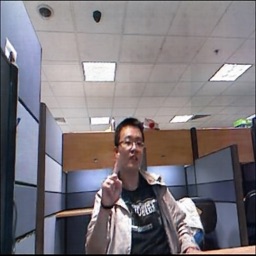}}\hfill
	\subfloat{%
		\includegraphics[width=\qwidth\textwidth]{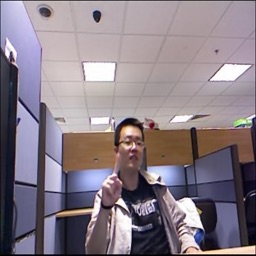}}
	\\
	\subfloat{%
		\includegraphics[width=\qwidth\textwidth]{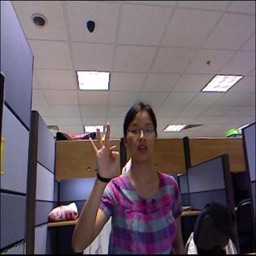}}\hfill
	\subfloat{%
		\includegraphics[width=\qwidth\textwidth]{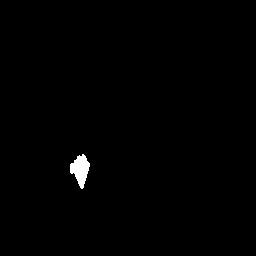}}\hfill
	\subfloat{%
		\includegraphics[width=\qwidth\textwidth]{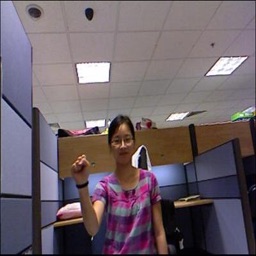}}\hfill
	\subfloat{%
		\includegraphics[width=\qwidth\textwidth]{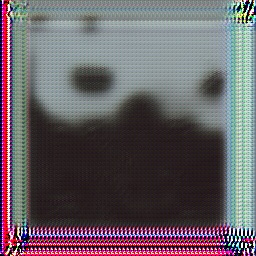}}\hfill
	\subfloat{%
		\includegraphics[width=\qwidth\textwidth]{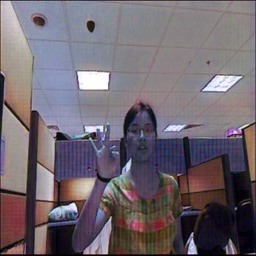}}\hfill
	\subfloat{%
		\includegraphics[width=\qwidth\textwidth]{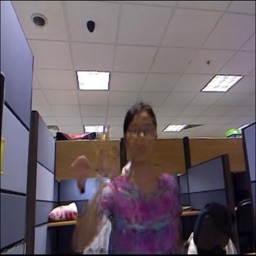}}\hfill
	\subfloat{%
		\includegraphics[width=\qwidth\textwidth]{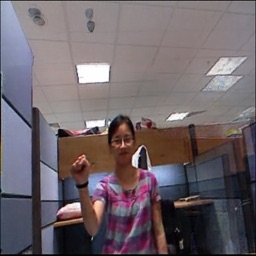}}\hfill
	\subfloat{%
		\includegraphics[width=\qwidth\textwidth]{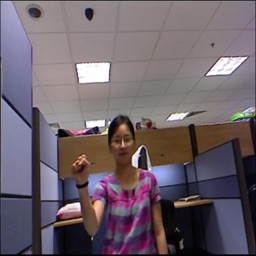}}
	\\
	\subfloat{%
		\includegraphics[width=\qwidth\textwidth]{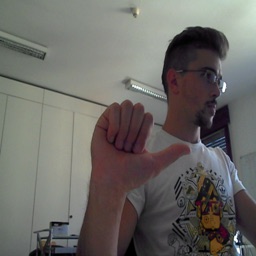}}\hfill
	\subfloat{%
		\includegraphics[width=\qwidth\textwidth]{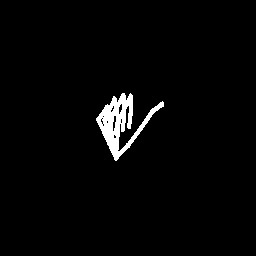}}\hfill
	\subfloat{%
		\includegraphics[width=\qwidth\textwidth]{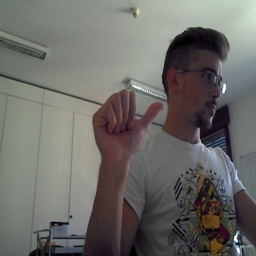}}\hfill
	\subfloat{%
		\includegraphics[width=\qwidth\textwidth]{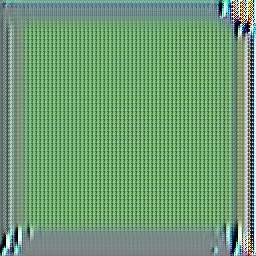}}\hfill
	\subfloat{%
		\includegraphics[width=\qwidth\textwidth]{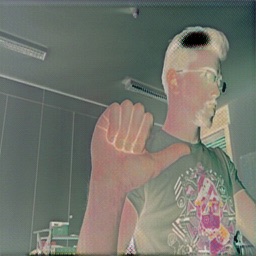}}\hfill
	\subfloat{%
		\includegraphics[width=\qwidth\textwidth]{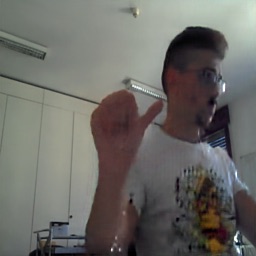}}\hfill
	\subfloat{%
		\includegraphics[width=\qwidth\textwidth]{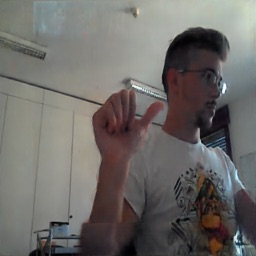}}\hfill
	\subfloat{%
		\includegraphics[width=\qwidth\textwidth]{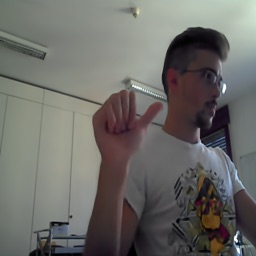}}
	\\
	\subfloat{%
		\includegraphics[width=\qwidth\textwidth]{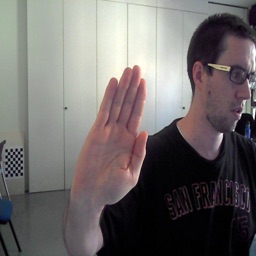}}\hfill
	\subfloat{%
		\includegraphics[width=\qwidth\textwidth]{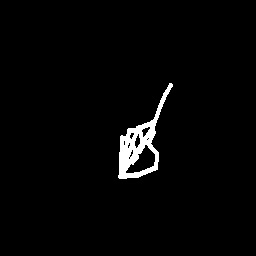}}\hfill
	\subfloat{%
		\includegraphics[width=\qwidth\textwidth]{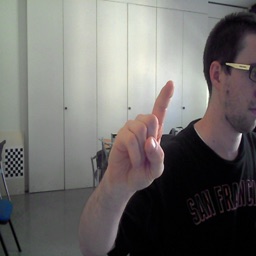}}\hfill
	\subfloat{%
		\includegraphics[width=\qwidth\textwidth]{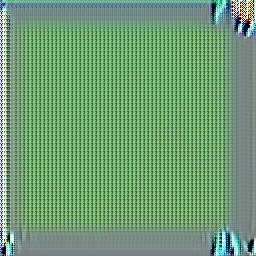}}\hfill
	\subfloat{%
		\includegraphics[width=\qwidth\textwidth]{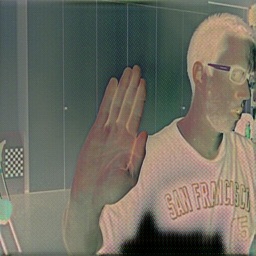}}\hfill
	\subfloat{%
		\includegraphics[width=\qwidth\textwidth]{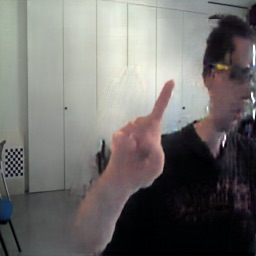}}\hfill
	\subfloat{%
		\includegraphics[width=\qwidth\textwidth]{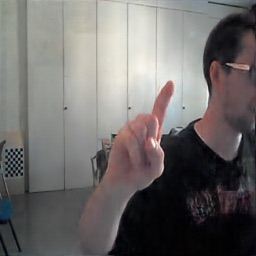}}\hfill
	\subfloat{%
		\includegraphics[width=\qwidth\textwidth]{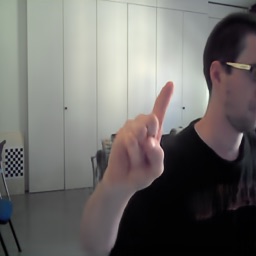}}
	\\
	\subfloat{%
		\includegraphics[width=\qwidth\textwidth]{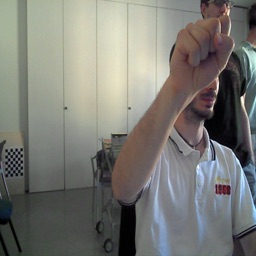}}\hfill
	\subfloat{%
		\includegraphics[width=\qwidth\textwidth]{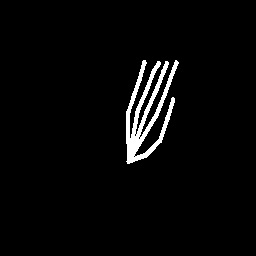}}\hfill
	\subfloat{%
		\includegraphics[width=\qwidth\textwidth]{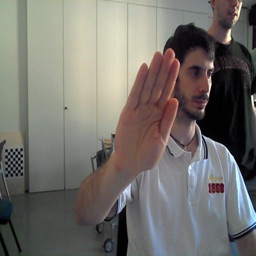}}\hfill
	\subfloat{%
		\includegraphics[width=\qwidth\textwidth]{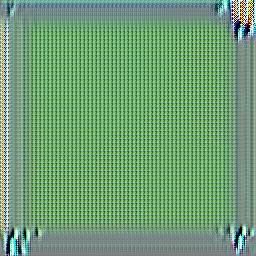}}\hfill
	\subfloat{%
		\includegraphics[width=\qwidth\textwidth]{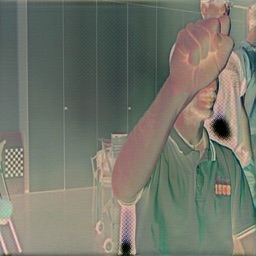}}\hfill
	\subfloat{%
		\includegraphics[width=\qwidth\textwidth]{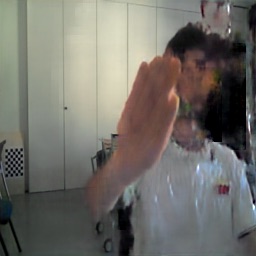}}\hfill
	\subfloat{%
		\includegraphics[width=\qwidth\textwidth]{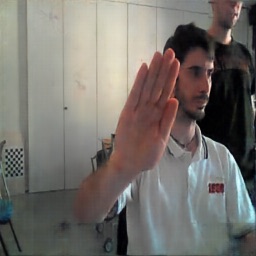}}\hfill
	\subfloat{%
		\includegraphics[width=\qwidth\textwidth]{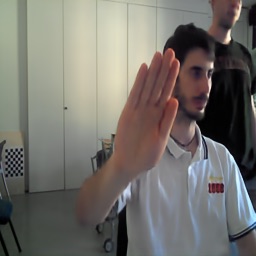}}
	\\
	\subfloat{%
		\includegraphics[width=\qwidth\textwidth]{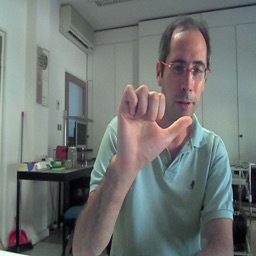}}\hfill
	\subfloat{%
		\includegraphics[width=\qwidth\textwidth]{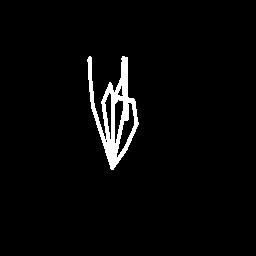}}\hfill
	\subfloat{%
		\includegraphics[width=\qwidth\textwidth]{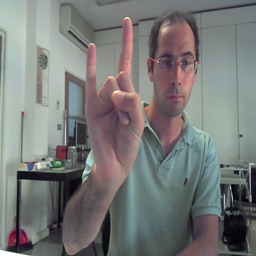}}\hfill
	\subfloat{%
		\includegraphics[width=\qwidth\textwidth]{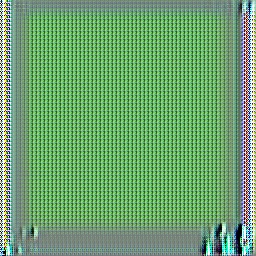}}\hfill
	\subfloat{%
		\includegraphics[width=\qwidth\textwidth]{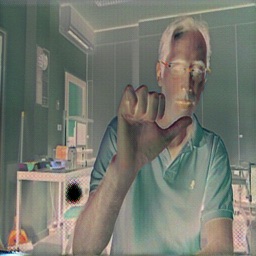}}\hfill
	\subfloat{%
		\includegraphics[width=\qwidth\textwidth]{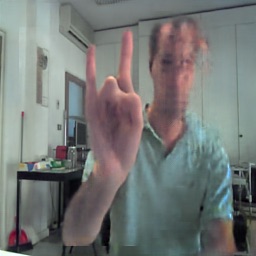}}\hfill
	\subfloat{%
		\includegraphics[width=\qwidth\textwidth]{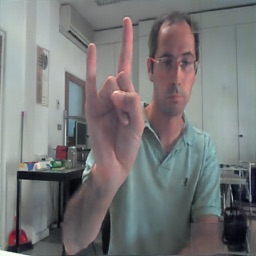}}\hfill
	\subfloat{%
		\includegraphics[width=\qwidth\textwidth]{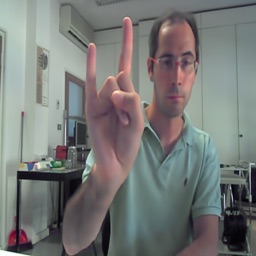}}
	
	\caption{Qualitative comparison between \TGAN~and competing works in the \textit{normal} experimental setting. NTU Hand Gesture dataset (top four rows) and Creative Senz3D (bottom four rows).}
	\label{fig:add_normal}
\end{figure*}

\begin{figure*}[!htp]
    \renewcommand{\tabcolsep}{1pt}
	\centering
	\subfloat[Source]{%
		\includegraphics[width=\qwidthL\textwidth]{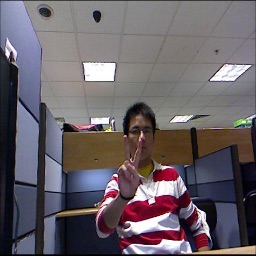}}\ 
	\subfloat[Target triangle]{%
		\includegraphics[width=\qwidthL\textwidth]{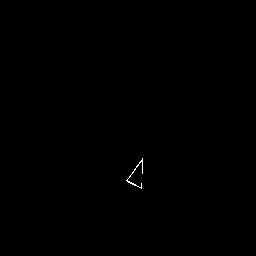}}\ 
	\subfloat[Target category]{%
		\squarecat{"8"}}\ 
	\subfloat[Ground truth]{%
		\includegraphics[width=\qwidthL\textwidth]{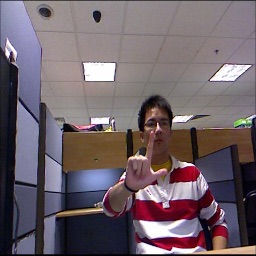}}\ 
	\subfloat[GestureGAN~\cite{tang2018gesturegan}]{%
		\includegraphics[width=\qwidthL\textwidth]{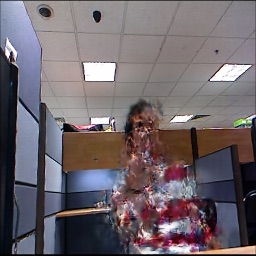}}\ 
	\subfloat[\boldTGAN ]{%
		\includegraphics[width=\qwidthL\textwidth]{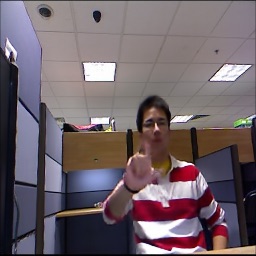}}
	\\
	\subfloat{%
		\includegraphics[width=\qwidthL\textwidth]{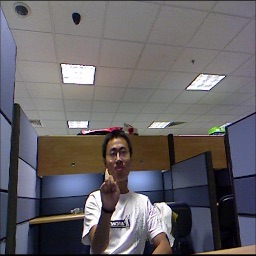}}\ 
	\subfloat{%
		\includegraphics[width=\qwidthL\textwidth]{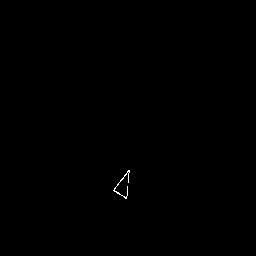}}\ 
	\subfloat{%
		\squarecat{"6"}}\ 
	\subfloat{%
		\includegraphics[width=\qwidthL\textwidth]{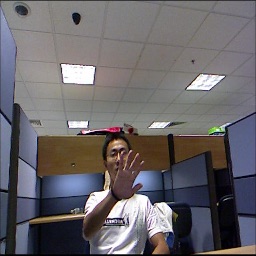}}\ 
	\subfloat{%
		\includegraphics[width=\qwidthL\textwidth]{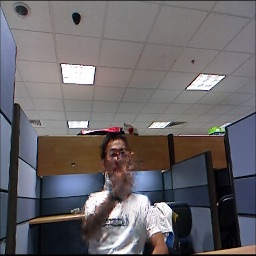}}\ 
	\subfloat{%
		\includegraphics[width=\qwidthL\textwidth]{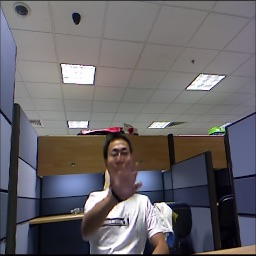}}
	\\
	\subfloat{%
		\includegraphics[width=\qwidthL\textwidth]{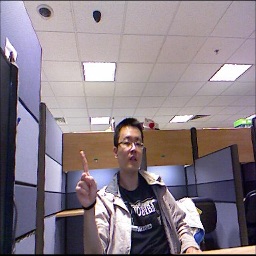}}\ 
	\subfloat{%
		\includegraphics[width=\qwidthL\textwidth]{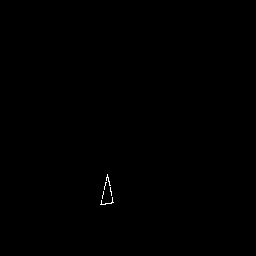}}\ 
	\subfloat{%
		\squarecat{"4"}}\ 
	\subfloat{%
		\includegraphics[width=\qwidthL\textwidth]{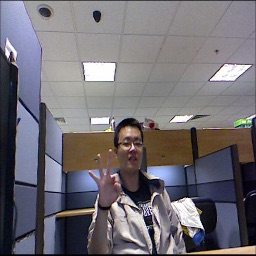}}\ 
	\subfloat{%
		\includegraphics[width=\qwidthL\textwidth]{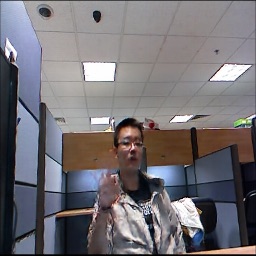}}\ 
	\subfloat{%
		\includegraphics[width=\qwidthL\textwidth]{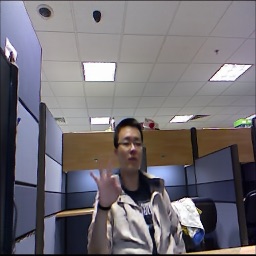}}
	\\
	\subfloat{%
		\includegraphics[width=\qwidthL\textwidth]{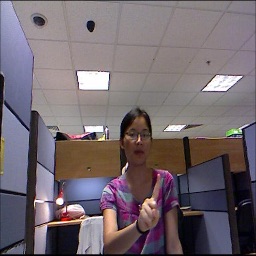}}\ 
	\subfloat{%
		\includegraphics[width=\qwidthL\textwidth]{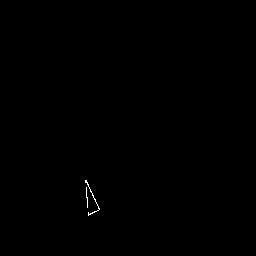}}\ 
	\subfloat{%
		\squarecat{"5"}}\ 
	\subfloat{%
		\includegraphics[width=\qwidthL\textwidth]{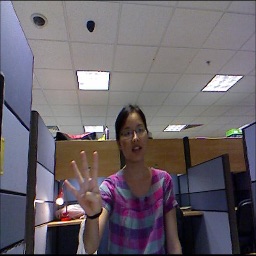}}\ 
	\subfloat{%
		\includegraphics[width=\qwidthL\textwidth]{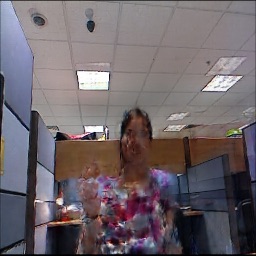}}\ 
	\subfloat{%
		\includegraphics[width=\qwidthL\textwidth]{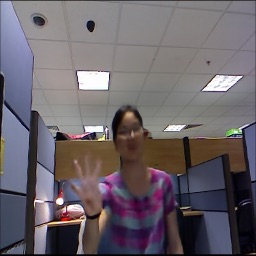}}
	\\
	\subfloat{%
		\includegraphics[width=\qwidthL\textwidth]{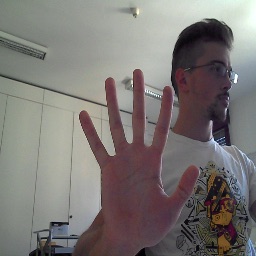}}\ 
	\subfloat{%
		\includegraphics[width=\qwidthL\textwidth]{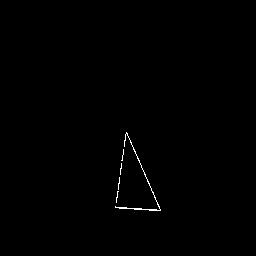}}\ 
	\subfloat{%
		\squarecat{"4"}}\ 
	\subfloat{%
		\includegraphics[width=\qwidthL\textwidth]{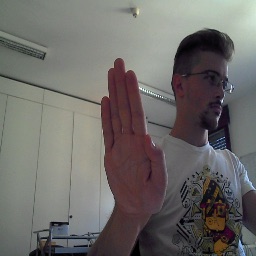}}\ 
	\subfloat{%
		\includegraphics[width=\qwidthL\textwidth]{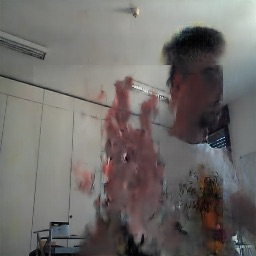}}\ 
	\subfloat{%
		\includegraphics[width=\qwidthL\textwidth]{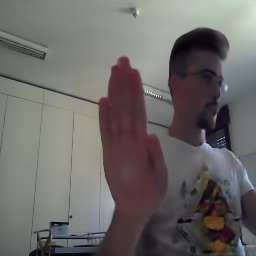}}
	\\
	\subfloat{%
		\includegraphics[width=\qwidthL\textwidth]{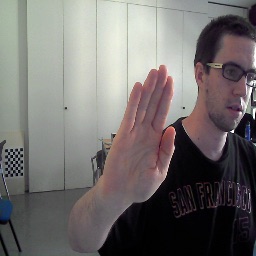}}\ 
	\subfloat{%
		\includegraphics[width=\qwidthL\textwidth]{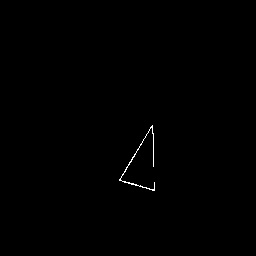}}\ 
	\subfloat{%
		\squarecat{"11"}}\ 
	\subfloat{%
		\includegraphics[width=\qwidthL\textwidth]{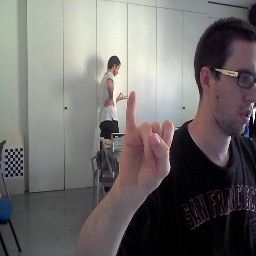}}\ 
	\subfloat{%
		\includegraphics[width=\qwidthL\textwidth]{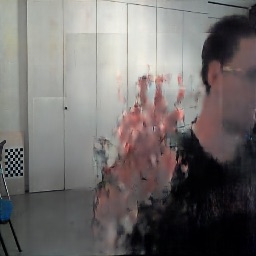}}\ 
	\subfloat{%
		\includegraphics[width=\qwidthL\textwidth]{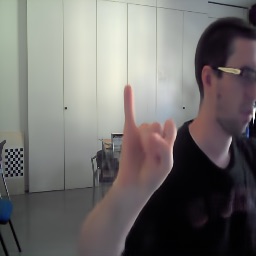}}
	\\
	\subfloat{%
		\includegraphics[width=\qwidthL\textwidth]{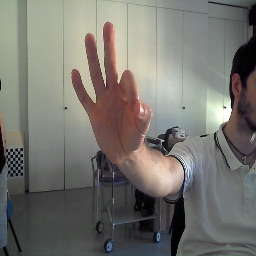}}\ 
	\subfloat{%
		\includegraphics[width=\qwidthL\textwidth]{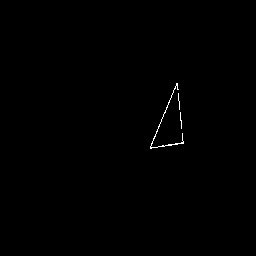}}\ 
	\subfloat{%
		\squarecat{"11"}}\ 
	\subfloat{%
		\includegraphics[width=\qwidthL\textwidth]{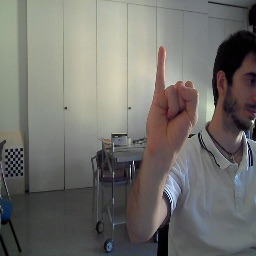}}\ 
	\subfloat{%
		\includegraphics[width=\qwidthL\textwidth]{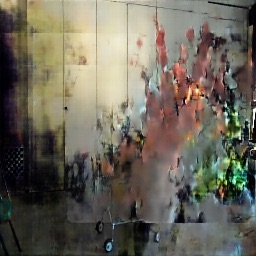}}\ 
	\subfloat{%
		\includegraphics[width=\qwidthL\textwidth]{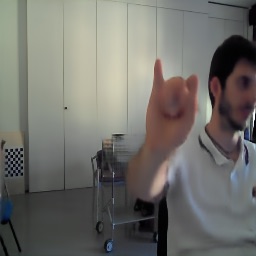}}
	\\
	\subfloat{%
		\includegraphics[width=\qwidthL\textwidth]{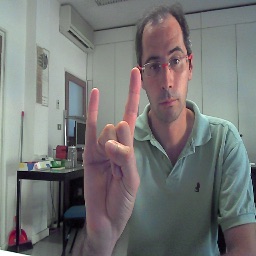}}\ 
	\subfloat{%
		\includegraphics[width=\qwidthL\textwidth]{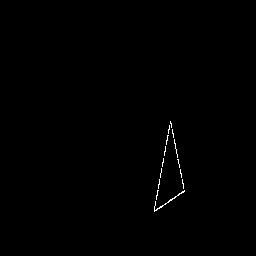}}\ 
	\subfloat{%
		\squarecat{"7"}}\ 
	\subfloat{%
		\includegraphics[width=\qwidthL\textwidth]{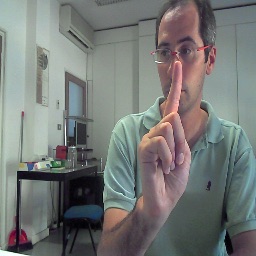}}\ 
	\subfloat{%
		\includegraphics[width=\qwidthL\textwidth]{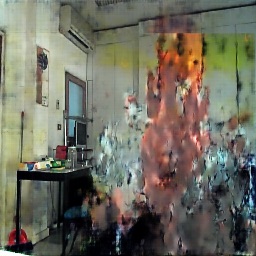}}\ 
	\subfloat{%
		\includegraphics[width=\qwidthL\textwidth]{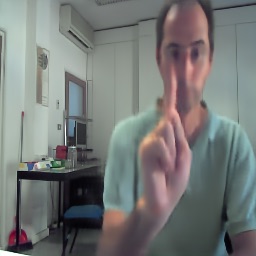}}
	
	\caption{Qualitative comparison between \TGAN~and competing works in the \textit{challenging} experimental setting. NTU Hand Gesture dataset (top four rows) and Creative Senz3D (bottom four rows).}
	\label{fig:add_challenging}
\end{figure*}

Figure~\ref{fig:add_normal} and Figure~\ref{fig:add_challenging} show additional images with 256$\times$256 resolutions generated by \TGAN\ in both \textit{normal} and \textit{challenging} settings, respectively.

\subsection{Feature Visualization}

\begin{figure*}[!htp]
    \renewcommand{\tabcolsep}{1pt}
	\centering
	\subfloat[Conditional map]{%
		\includegraphics[width=\qwidthL\textwidth]{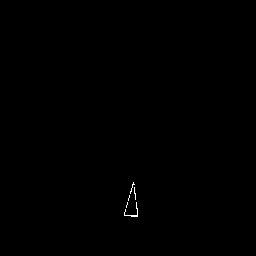}}\ 
	\subfloat[Ground truth]{%
		\includegraphics[width=\qwidthL\textwidth]{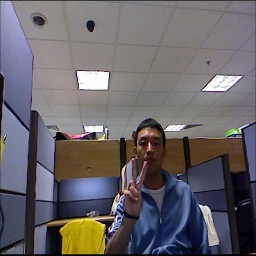}}\ 
	\subfloat[NO rolling]{%
		\includegraphics[width=\qwidthL\textwidth]{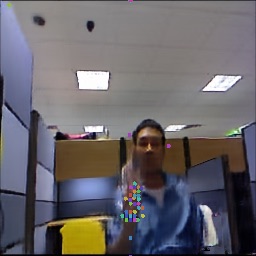}}\ 
	\subfloat[WITH rolling]{%
		\includegraphics[width=\qwidthL\textwidth]{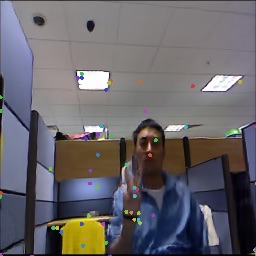}}
	\\
	\subfloat{%
		\includegraphics[width=\qwidthL\textwidth]{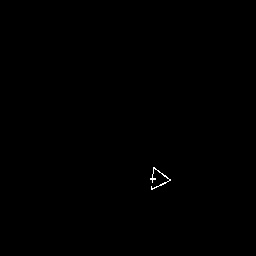}}\ 
	\subfloat{%
		\includegraphics[width=\qwidthL\textwidth]{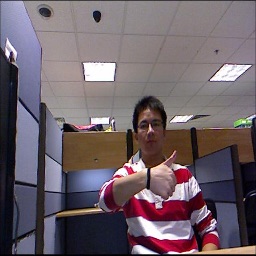}}\ 
	\subfloat{%
		\includegraphics[width=\qwidthL\textwidth]{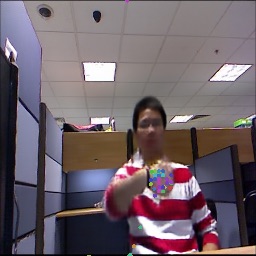}}\ 
	\subfloat{%
		\includegraphics[width=\qwidthL\textwidth]{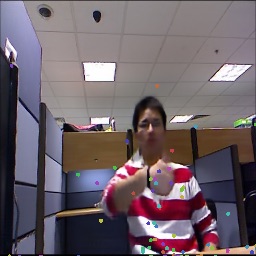}}
	\\
	\subfloat{%
		\includegraphics[width=\qwidthL\textwidth]{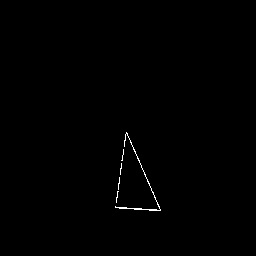}}\ 
	\subfloat{%
		\includegraphics[width=\qwidthL\textwidth]{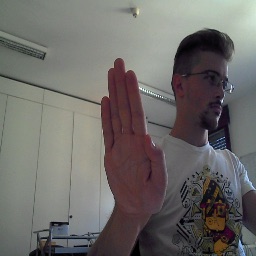}}\ 
	\subfloat{%
		\includegraphics[width=\qwidthL\textwidth]{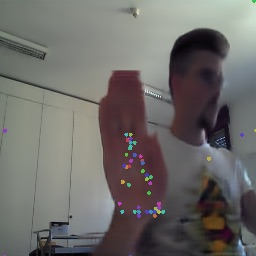}}\ 
	\subfloat{%
		\includegraphics[width=\qwidthL\textwidth]{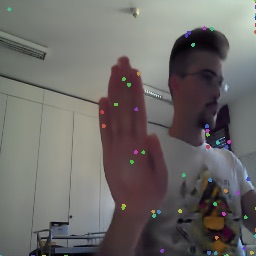}}
	\\
	\subfloat{%
		\includegraphics[width=\qwidthL\textwidth]{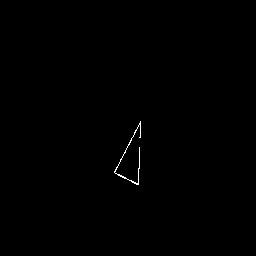}}\ 
	\subfloat{%
		\includegraphics[width=\qwidthL\textwidth]{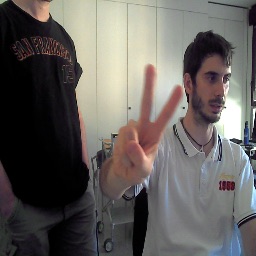}}\ 
	\subfloat{%
		\includegraphics[width=\qwidthL\textwidth]{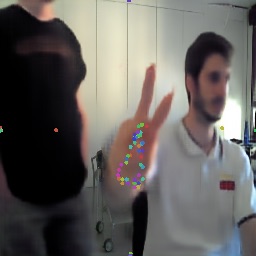}}\ 
	\subfloat{%
		\includegraphics[width=\qwidthL\textwidth]{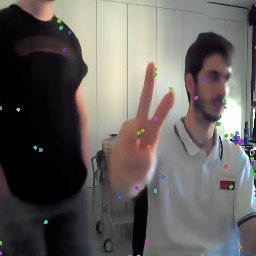}}
	\\
	\subfloat{%
		\includegraphics[width=\qwidthL\textwidth]{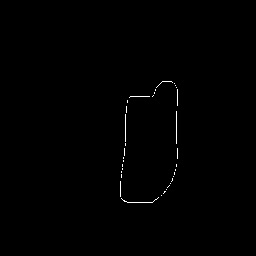}}\ 
	\subfloat{%
		\includegraphics[width=\qwidthL\textwidth]{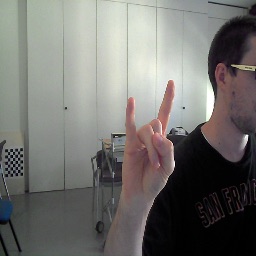}}\ 
	\subfloat{%
		\includegraphics[width=\qwidthL\textwidth]{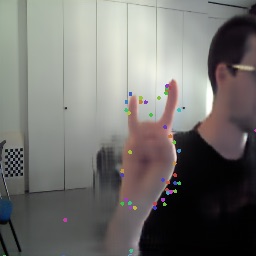}}\ 
	\subfloat{%
		\includegraphics[width=\qwidthL\textwidth]{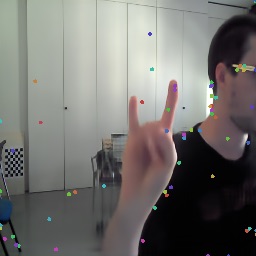}}
	\\ 
	\subfloat{%
		\includegraphics[width=\qwidthL\textwidth]{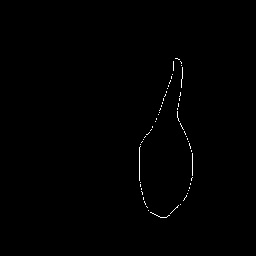}}\ 
	\subfloat{%
		\includegraphics[width=\qwidthL\textwidth]{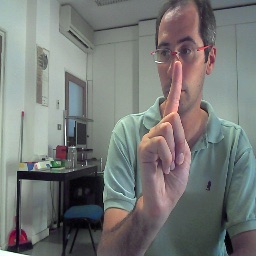}}\ 
	\subfloat{%
		\includegraphics[width=\qwidthL\textwidth]{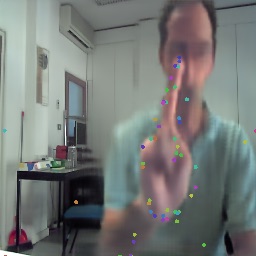}}\ 
	\subfloat{%
		\includegraphics[width=\qwidthL\textwidth]{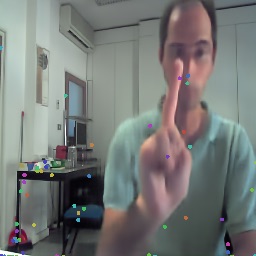}}
	\\ 
	\subfloat{%
		\includegraphics[width=\qwidthL\textwidth]{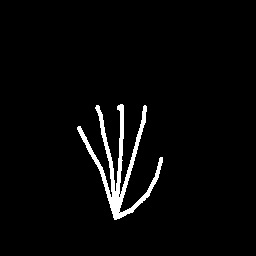}}\ 
	\subfloat{%
		\includegraphics[width=\qwidthL\textwidth]{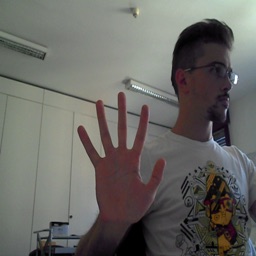}}\ 
	\subfloat{%
		\includegraphics[width=\qwidthL\textwidth]{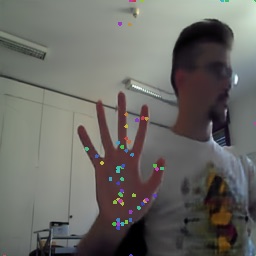}}\ 
	\subfloat{%
		\includegraphics[width=\qwidthL\textwidth]{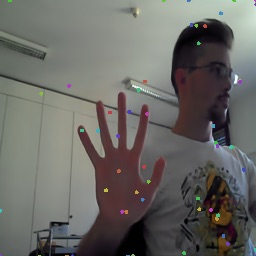}}
	\\
	\subfloat{%
		\includegraphics[width=\qwidthL\textwidth]{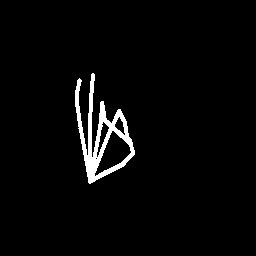}}\ 
	\subfloat{%
		\includegraphics[width=\qwidthL\textwidth]{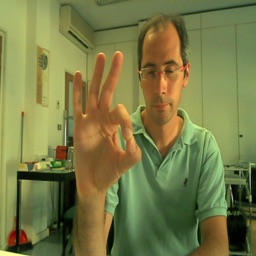}}\ 
	\subfloat{%
		\includegraphics[width=\qwidthL\textwidth]{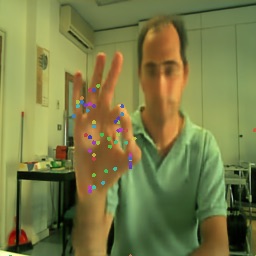}}\ 
	\subfloat{%
		\includegraphics[width=\qwidthL\textwidth]{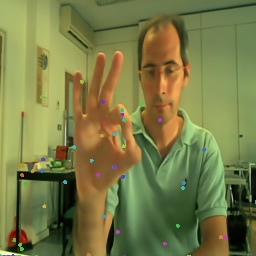}}
	\\
	
	\caption{Visualization of the encoded features with different conditional map shapes: triangle, boundary and skeleton. It is possible to see that the network, before rolling, focuses mostly on the gesture. Rolling allows to focus to the background context as well, thus increasing the quality of the image.}
	\label{fig:visualization}
\end{figure*}

Inspired by Monkey-Net~\cite{siarohin2018animating}, we employ softmax activations for the last layer of the encoder $E_2$ to obtain heatmaps, which transfers the $K$-channels feature maps into heatmaps $H_K\in [0, 1.0]^{\frac{H}{4}\times\frac{W}{4}}$, one for each channel. Thus, we estimate the location of the confidence from the heatmaps, which can reveal the spatial distribution of learned features.
Figure~\ref{fig:visualization} shows that the rolling guidance approach helps the model to learn additional information that significantly refines the generated results.

\subsection{Failure Cases}
\begin{figure*}[!htp]
    \renewcommand{\tabcolsep}{1pt}
	\centering
	\subfloat{%
		\squarecat{(a)}}\ 
	\subfloat[Source]{%
		\includegraphics[width=\qwidthL\textwidth]{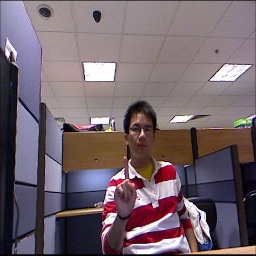}}\ 
	\subfloat[Conditional map]{%
		\includegraphics[width=\qwidthL\textwidth]{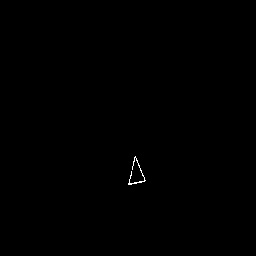}}\ 
	\subfloat[Ground Truth]{%
		\includegraphics[width=\qwidthL\textwidth]{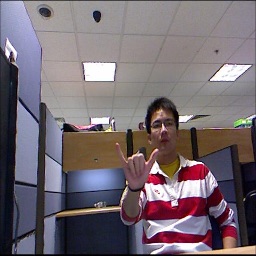}}\ 
	\subfloat[Attention mask]{%
		\includegraphics[width=\qwidthL\textwidth]{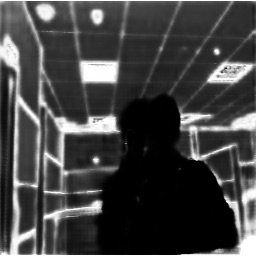}}\
	\subfloat[\TGAN]{%
		\includegraphics[width=\qwidthL\textwidth]{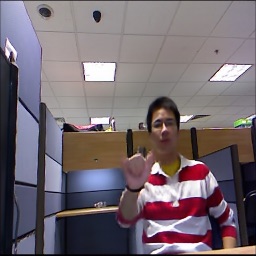}}
	\\
	\subfloat{%
		\squarecat{(b)}}\ 
	\subfloat{%
		\includegraphics[width=\qwidthL\textwidth]{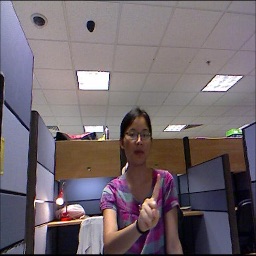}}\ 
	\subfloat{%
		\includegraphics[width=\qwidthL\textwidth]{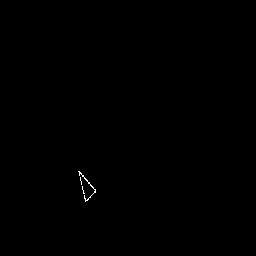}}\ 
	\subfloat{%
		\includegraphics[width=\qwidthL\textwidth]{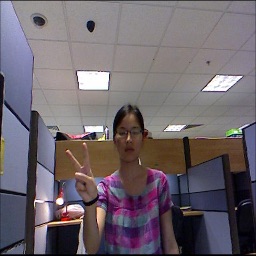}}\ 
	\subfloat{%
		\includegraphics[width=\qwidthL\textwidth]{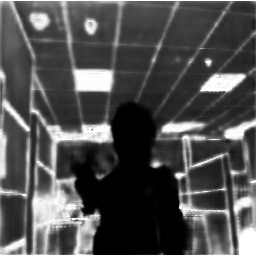}}\
	\subfloat{%
		\includegraphics[width=\qwidthL\textwidth]{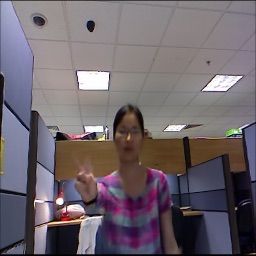}}
	\\
	\subfloat{%
		\squarecat{(c)}}\ 
	\subfloat{%
		\includegraphics[width=\qwidthL\textwidth]{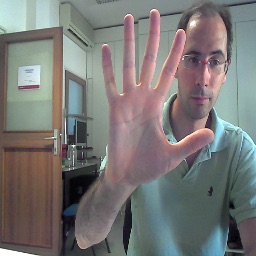}}\ 
	\subfloat{%
		\includegraphics[width=\qwidthL\textwidth]{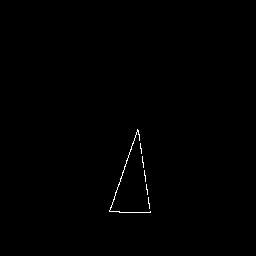}}\ 
	\subfloat{%
		\includegraphics[width=\qwidthL\textwidth]{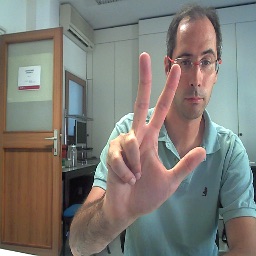}}\ 
	\subfloat{%
		\includegraphics[width=\qwidthL\textwidth]{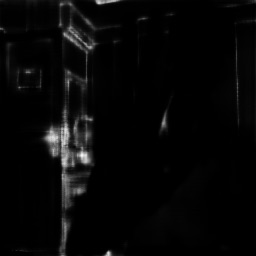}}\
	\subfloat{%
		\includegraphics[width=\qwidthL\textwidth]{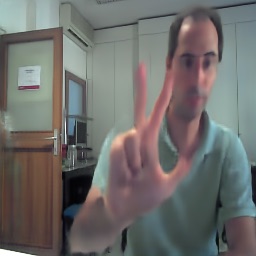}}
	\\
	\subfloat{%
		\squarecat{(d)}}\ 
	\subfloat{%
		\includegraphics[width=\qwidthL\textwidth]{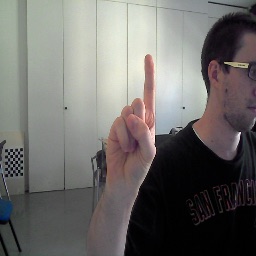}}\ 
	\subfloat{%
		\includegraphics[width=\qwidthL\textwidth]{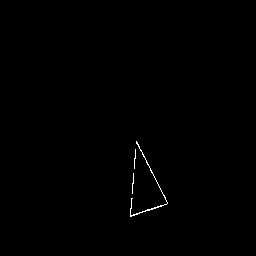}}\ 
	\subfloat{%
		\includegraphics[width=\qwidthL\textwidth]{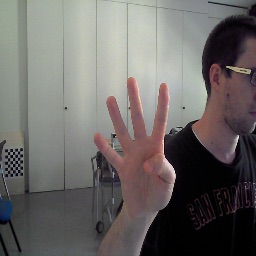}}\ 
	\subfloat{%
		\includegraphics[width=\qwidthL\textwidth]{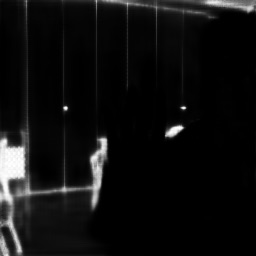}}\
	\subfloat{%
		\includegraphics[width=\qwidthL\textwidth]{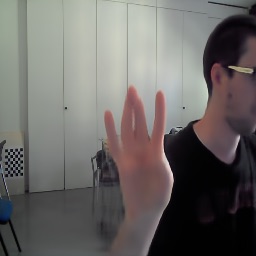}}
	
	\caption{Some examples of failure cases of our method.}
	\label{fig:failures}
\end{figure*}

Figure~\ref{fig:failures} shows sample failure cases of our method. (a), (b) and (c) show that our method may lose some details of finger or arm sometimes, especially in the NTU dataset, in which hand gestures are very small. (d) shows some distortion of the generated gesture. Handling more diverse and extreme transformations, especially geometric constraints, is an important problem for future work.

\end{document}